\documentclass[sigconf, screen]{acmart}
\AtBeginDocument{%
  }

\setcopyright{acmlicensed}
\copyrightyear{2018}
\acmYear{2018}
\acmDOI{XXXXXXX.XXXXXXX}
\acmConference[Conference acronym 'XX]{Make sure to enter the correct
 conference title from your rights confirmation email}{June 03--05,
 2018}{Woodstock, NY}
\acmISBN{978-1-4503-XXXX-X/2018/06}

\usepackage{multirow}
\usepackage{subcaption}
\usepackage{booktabs}
\usepackage{url}
\usepackage{color}
\usepackage{algorithm}
\usepackage{enumitem}
\usepackage{graphicx}
\usepackage[noend]{algpseudocode}
\usepackage{hyperref}
\hypersetup{
    colorlinks=true,
    linkcolor=blue,
    filecolor=magenta,      
    urlcolor=blue,
    pdftitle={main},
    pdfpagemode=FullScreen,
    }
\usepackage{dblfloatfix}
\usepackage{afterpage}

\makeatletter
\def\algbackskip{\hskip-\ALG@thistlm}
\makeatother

\newcommand{\method}[1]{\textsf{Method}}

\begin{document}

\title{FairExpand: Individual Fairness on Graphs 
with~Partial~Similarity~Information}

\author{Rebecca Salganik}
\email{rsalgani@ur.rochester.edu}
\orcid{0009-0007-9273-8780}
\affiliation{%
  \institution{University of Rochester}
  \city{Rochester}
  \state{NY}
  \country{USA}
}
\author{Guillaume Salha-Galvan}
\email{gsalhagalvan@gmail.com}
\orcid{0000-0002-2452-1041}
\affiliation{%
  \institution{SJTU Paris Elite Institute of Technology
}
  \city{Shanghai}
  \country{China}
}
\author{Yibin Wang}
\email{ywang448@ur.rochester.edu}
\affiliation{%
  \institution{University of Rochester}
  \city{Rochester}
  \state{NY}
  \country{USA}
}

\author{Jian Kang}
\email{jian.kang@mbzuai.ac.ae}
\orcid{0000-0003-3902-7131}
\affiliation{%
  \institution{MBZUAI}
  \city{Abu Dhabi}
  \country{UAE}
}

\begin{abstract}
   
Individual fairness, which requires that similar individuals should be treated similarly by algorithmic systems, has become a central principle in fair machine learning. Individual fairness has garnered traction in graph representation learning due to its practical importance in high-stakes Web areas such as user modeling, recommender systems, and search. However, existing methods assume the existence of predefined similarity information over all node pairs, an often unrealistic requirement that prevents their operationalization in practice.
In this paper, we assume the similarity information is only available for a limited subset of node pairs and introduce FairExpand, a flexible framework that promotes individual fairness in this more realistic partial information scenario. FairExpand follows a two-step pipeline that alternates between refining node representations using a backbone model (e.g., a graph neural network) and gradually propagating similarity information, which allows fairness enforcement to effectively expand to the entire graph. Extensive experiments show that FairExpand consistently enhances individual fairness while preserving performance, making it a practical solution for enabling graph-based individual fairness in real-world applications with partial similarity information.

\end{abstract}



\maketitle

% \vspace{-1em}
\section{Introduction}

Fairness has become a prominent area of research in machine learning over recent years \cite{pessach2022review,caton2024fairness,mehrabi2021survey}. As algorithmic systems increasingly shape decisions in high-stakes domains, including Web applications such as social media personalization and cultural content recommendation, as well as broad sectors such as finance, healthcare, and hiring, debates about their responsibilities have intensified. This renewed attention reflects growing concerns about the risks of bias and discrimination that may disproportionately affect vulnerable or marginalized users \cite{waddel_how_2016,andrews_how_2021,chen_ethics_2023,salganik_music_2024,caton2024fairness,ferraro_what_2021,dowrk_fairness_2012}. To tackle these challenges, various definitions of fairness have been proposed. Among these, one prominent formalization is \textit{individual fairness} \cite{dowrk_fairness_2012}, which hinges on the premise that similar individuals should be treated similarly by an algorithmic system~\cite{dwork2020individual,gupta2021individual,zemel2013learning,lahoti2019ifair}. This principle offers a personalized lens on fairness, distinguishing it from group-level notions that often rely on demographic attribute(s) \cite{barocas2023fairness,binns2020apparent}.

Promoting individual fairness is particularly crucial in \textit{graph representation learning}, which focuses on learning node embedding vectors, i.e., node representations, from the structural and attributive information of a graph \cite{hamilton2020graph,ju2024comprehensive,kipf_semi_2016,velivckovic2023everything}. For example, in content recommendation, a graph-based model may encode biases from both a user-item interaction graph and the individual node features, creating disparities in the representations and treatment of similar users or creators, causing poor performance or obscuring relevant content~\cite{salganik_fairness_2024,marras_equality_2022}.

Several graph-based individual fairness methods have been proposed to ensure that nodes deemed similar according to some measure are mapped to representations that remain similar in the embedding space \cite{kang_inform_2020,dong_individual_2021,xu_gfairhint_2023,chen2024fairness,wang2024individual,laclau2022survey}. Intuitively, this enforced alignment ensures that similar nodes yield consistent predictions in downstream tasks \cite{kang_inform_2020,dong_individual_2021}. Most existing approaches compare some form of apriori similarity information with a model's output and penalize deviations between the two, using weighted-sum losses \cite{kang_inform_2020}, ranking losses \cite{dong_individual_2021}, or other related mechanisms \cite{zhu_devil_2024}.

However, a critical bottleneck in these existing methods is their reliance on a predefined similarity measure \textit{assumed available for all node pairs in the graph}. In practice, this assumption is often unrealistic for two reasons: (1) calculating complete pairwise similarities on large graphs is computationally demanding \cite{hamilton2017representation}, or (2) obtaining similarity annotations may require costly expert annotations \cite{dowrk_fairness_2012,kim_fairness_2018}. In  many industry settings, including million-scale Web applications, these scalability limitations pose serious obstacles to the operationalization of individual fairness in practice.

In this paper, we address this challenge by introducing FairExpand, a flexible framework where similarity guidance is provided only for a limited subset of nodes. FairExpand follows an iterative pipeline with two components: (1)~a~backbone model that learns representations while enforcing partial fairness from the available similarity data, and (2) an auxiliary link prediction model that interpolates between exploration and exploitation to propagate similarity information across the graph. By alternating between representation refinement and similarity expansion, FairExpand scales fairness promotion to the entire graph. Since obtaining  limited annotations is substantially more feasible, FairExpand aims to enable the practical adoption of graph-based individual fairness, thereby bridging the gap between individual fairness theory and real-world applications. In summary, our contributions are as follows:

\begin{enumerate}
\item We present a formalization of the individual fairness problem in graph representation learning under partial similarity information, emphasizing the key components and constraints of this more realistic setting.

\item We introduce FairExpand, a flexible framework for learning individually fair node representations with partial similarity information. To our knowledge, this is the first approach for effectively enforcing individual fairness on graphs in the more realistic scenario of partial similarity information.

\item We evaluate the empirical effectiveness of FairExpand via comprehensive node classification experiments on six real-world graphs. Our framework consistently enhances individual fairness, even when similarity information is limited, while maintaining high classification~accuracy.

\item We release our source code\footnote{\url{https://anonymous.4open.science/r/FairExpand-WWW-47BC/README.md}} to facilitate the future use of FairExpand and ensure the reproducibility of our~results.

\end{enumerate}

% This paper is organized as follows.
% Section~\ref{sec:related-work} reviews the relevant background and related work.
% Section~\ref{sec:problemformulation} then formalizes our problem.
% Section~\ref{section4} presents the FairExpand framework.
% Section~\ref{sec:experiments} reports our experimental evaluation, and Section~\ref{conclusion} concludes.

\section{Background and Related Work}\label{sec:related-work}
Fairness in machine learning has received substantial attention \cite{mehrabi2021survey, pessach2022review, caton2024fairness}. This section reviews prior work on individual fairness, individual fairness on graphs, and fairness with partial similarity, which together form the background of our contribution.

\subsection{Individual Fairness}

\citet{dowrk_fairness_2012} introduced the principle of individual fairness, which states that \textit{similar individuals} should be treated similarly by algorithmic systems. Formally, let $x_i$ denote individuals, and let $D$ be a \textit{distance measure}, interpreted as a proxy for similarity. A model $f$ is said to be individually fair if:
\begin{equation}
D'\big(f(x_i), f(x_j)\big) \leq L \times D(x_i, x_j), \quad \forall (x_i, x_j),
\label{ifequation}
\end{equation}
for some output-space metric $D'$ and Lipschitz constant $L > 0$. This smoothness constraint enforces consistent model outcomes for similar individuals and contrasts with group fairness notions, which target statistical parity across  groups~\cite{mehrabi2021survey}. It provides a principled way to limit individual disparities and has inspired a substantial line of work across various domains such as healthcare~\cite{zemel2013learning}, recidivism prediction~\cite{lahoti2019operationalizing}, search engines~\cite{biega2018equity}, and recommender systems~\cite{li2022fairness}, as well as graph learning, which we discuss next.

\subsection{Individual Fairness on Graphs} \label{subsec:related-ifgraph}

Individual fairness on graphs extends the principle to relational data, requiring that nodes deemed similar in a graph, under a specified measure, receive similar predictions or representations from a model such as a graph neural network (GNN)~\cite{hamilton2020graph}. This setting introduces additional challenges, since both similarity information and relational structure must be modeled jointly.

\citet{kang_inform_2020} proposed the first individual fairness framework for graphs, InFoRM, which enforces constraints akin to Equation~\eqref{ifequation} by minimizing a Laplacian regularization term over the model's outputs and an a priori node-to-node similarity matrix. Later, \citet{dong_individual_2021} presented REDRESS, an alternative framework that models individual fairness from a ranking perspective to avoid the absolute distance comparisons of InFoRM. \citet{salganik2022analyzing} analyzed the effects of various sampling strategies on bias using REDRESS. \citet{song2022guide} proposed enforcing individual fairness in GNNs while ensuring that fairness levels are balanced across demographic groups. \citet{xu_gfairhint_2023} introduced GFairHint, which uses link prediction on a similarity graph to inject fairness hints into GNN-based node embedding representations. Several works have also extended the notion of individual fairness to dynamic~\cite{song_towards_2023} or multi-view graphs~\cite{wang_ifig_2022}, while \citet{h2_on_2024} analyzed the sensitivity of various metrics used for individual fairness on graphs. Finally, Petersen et al. \cite{petersen_post_2021} developed a graph-based post-processing algorithm to improve individual fairness on top of black-box model outputs.

Importantly, these studies all assume full knowledge of pairwise similarity information. In practice, this is unrealistic: computing similarities for all node pairs is computationally prohibitive for large graphs \cite{hamilton2017representation}, and collecting accurate similarity labels may  require costly expert annotations \cite{dowrk_fairness_2012, kim_fairness_2018}. These limitations pose barriers to the implementation of these methods in real-world applications.

\subsection{Relaxing the Full Information Assumption}
\label{subsec:related-partial}
Overall, even outside the graph domain, individual fairness studies still often assume full pairwise similarity information for all individuals. Several lines of work have begun to relax this assumption for non-graph data and highlight its  limitations. \citet{lahoti2019operationalizing} learned representations from partial similarity information obtained via pairwise fairness judgments for tabular data. \citet{kim_fairness_2018} studied a setting where the similarity measure can only be queried a limited number of times. Mukherjee et al. \cite{mukherjee_two_2020} inferred similarity measure from heterogeneous data modalities.  \citet{zhang_matrix_2023} showed that matrix estimation techniques can recover desirable individual fairness properties without full similarity information.

We note that relaxing full information requirements is also challenging \textit{beyond individual fairness}. In group and counterfactual fairness methods that typically require complete sensitive attributes, only a few studies attempt to address this limitation. Hashimoto et al.~\cite{tatsunori_fairness_2018} achieved group fairness without demographic attributes via distributionally robust optimization. Lahoti et al.~\cite{lahoti_fairness_2020} co-trained an adversary to identify regions where a model incurs large errors when sensitive attributes are unavailable. Dai et al.~\cite{dai_say_2021} inferred pseudo-sensitive labels from limited supervision using adversarial learning and covariance minimization.

\subsection{Limitations in Existing Works}
In summary, existing work reveals two key limitations. First, methods for individual fairness on graphs uniformly rely on full similarity information. Second, approaches that do relax full information requirements either require a costly expert query protocol~\cite{kim_fairness_2018}
or function by reconstructing a full similarity measure first, and then applying it to an input~\cite{mukherjee_two_2020, zhang_matrix_2023}. To our knowledge, FairExpand is the first framework for individual fairness on graphs with partial similarity information.

\section{Problem Formulation}
\label{sec:problemformulation}
In this section, we formalize the individual fairness problem on graphs with partial similarity information that we study in this~work.

\subsection{Problem Setting and Notation}
\label{setting}

\subsubsection{Graph Representation Learning}
We consider an undirected graph $\mathcal{G} = (\mathcal{V}, \mathcal{E})$ with a node set $\mathcal{V}$ of $n$ nodes and an edge set $\mathcal{E}$ of $m$ edges, represented by an $n \times n$ adjacency matrix $A$. Each node $i \in \mathcal{V}$ is described by a $d_1$-dimensional feature vector $x_i$, forming an $n \times d_1$ feature matrix $X$. 

We further consider a backbone graph representation learning model $\mathcal{M}$ that processes $(A, X)$ and produces a $d_2$-dimensional embedding vector $\hat{y}_i$ for each node $i \in \mathcal{V}$, with $d_2 \ll n$~\cite{hamilton2020graph}.~Formally:
\begin{equation}
\hat{Y} = \mathcal{M}(A, X),
\end{equation}
where $\hat{y}_i$ is the $i$-th row of the resulting $n \times d_2$ embedding matrix $\hat{Y}$. In what follows, we focus\footnote{Although we present FairExpand under this setting for clarity, we show throughout this paper that the framework is flexible and can accommodate other models and tasks.} on the case where $\mathcal{M}$ is a GNN trained for node classification~\cite{wu2020comprehensive,kipf_semi_2016}.

\subsubsection{Partial Similarity Information}
We assume the existence of a similarity measure over all nodes, represented by an $n \times n$ similarity matrix $S$ with entries $S_{ij} \geq 0$ quantifying the similarity between $(i, j) \in \mathcal{V}^2$. \textit{However, $S$ is not fully observable}. We only access the partial similarity matrix $S^{(0)}$, a masked version of~$S$~defined~as:
\begin{equation}
S^{(0)}_{ij} =
\begin{cases}
S_{ij}, & \text{if the similarity between $i$ and $j$ is observed,} \\
0, & \text{otherwise.}
\end{cases}
\end{equation}
We make no assumption about how $S$ is defined. It may rely on any criterion provided by domain experts, or computed through a chosen procedure. Likewise, we do not assume any particular mechanism for how $S^{(0)}$ is sampled from $S$, keeping our formulation general. 
Exact details of our implementation of $S$ and $S^{(0)}$ in experiments are provided in Section~\ref{sec:building_S}.

\subsection{Problem Objective}
Our goal is to ensure that the node representations learned by $\mathcal{M}$ are \emph{individually fair} with respect to the ground-truth similarity matrix $S$, while preserving node classification performance. Unlike prior work that assumes full access to $S$ during both training and testing, we consider a \emph{partial-information} setting: the model observes only the incomplete similarity matrix $S^{(0)}$ during training but is evaluated on fairness with respect to the entries of $S$ associated with the testing node set.

To quantify individual fairness, we adapt the InFoRM bias measure~\cite{kang_inform_2020} to this setting. Given learned node representations $\hat{Y}$, we define:
\begin{equation}\label{equationbias}
\text{Bias}(\hat{Y}, S)
= \sum_{i=1}^n \sum_{j=1}^n S_{ij}\,\|\hat{y}_i - \hat{y}_j\|_2^2
= \mathrm{Tr}(\hat{Y}^\top L_S \hat{Y}),
\end{equation}
where $L_S$ is the Laplacian of $S$~\cite{merris1994laplacian} and $\mathrm{Tr}(\cdot)$ denotes the trace operator~\cite{harville1997trace}. Higher values indicate that similar nodes receive inconsistent representations, while lower values correspond to greater individual fairness.

\section{Promoting Individual Fairness on Graphs with Partial Similarity Information}
\label{section4}

\begin{figure*}[]
    \centering
    
    \includegraphics[width=0.85\linewidth]{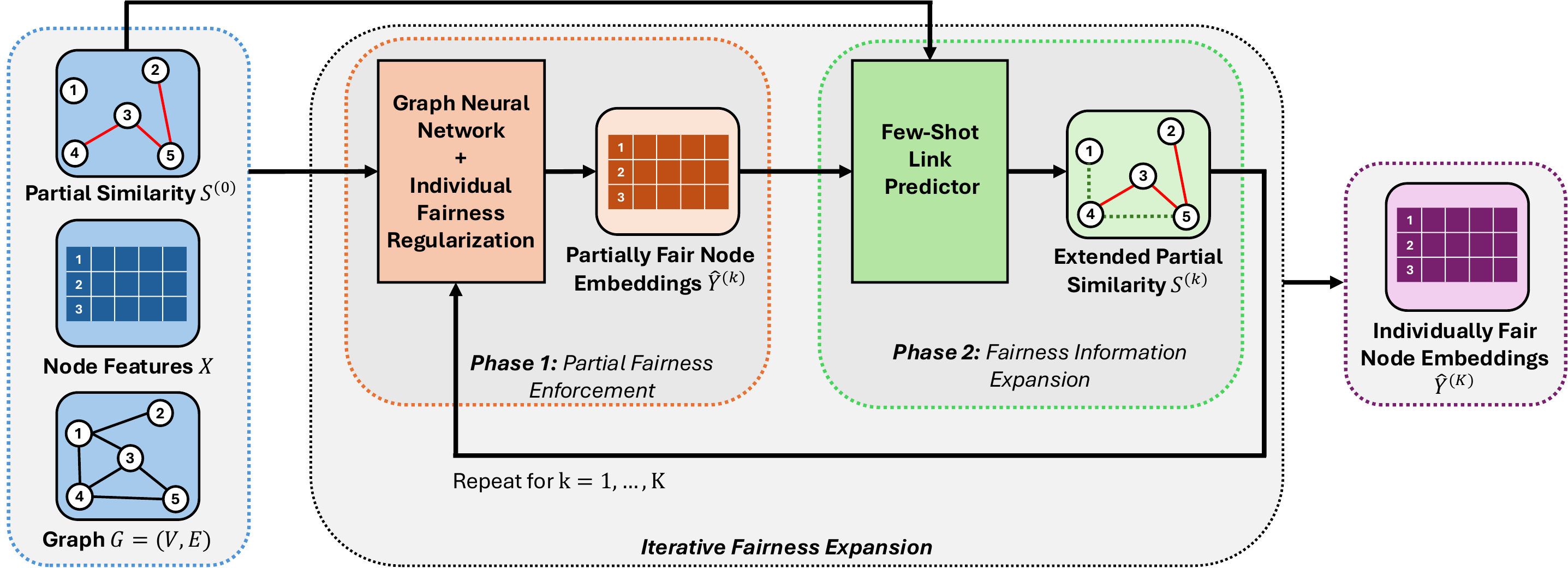}
  \caption{Illustration of FairExpand for individually fair graph representation learning with partial similarity information. This framework operates as an iterative pipeline with two components: (1) a partial fairness enforcement phase, where individual fairness is enforced on a backbone GNN model using the observed similarity data $S^{(0)}$, and (2) a similarity propagation phase, which uses an auxiliary link predictor and exploration techniques to gradually expand similarity information to larger regions of the graph. By alternating between these phases, FairExpand effectively scales fairness enforcement across the entire graph.}
    % \vspace{-4mm}
    \label{fig:framework}
\end{figure*}

This section introduces FairExpand. As illustrated in Figure~\ref{fig:framework}, this framework operates as an iterative pipeline with two phases:

\begin{enumerate}
\item \textbf{Phase 1:} a partial fairness enforcement phase, where the backbone model $\mathcal{M}$ under consideration learns node embedding representations while enforcing partial individual fairness based on the observed similarity data.
\item \textbf{Phase 2:} a similarity propagation phase, which uses an auxiliary link predictor and exploration techniques to gradually expand similarity information to larger parts of~the~graph.
\end{enumerate}

By alternating between refining representations and expanding similarity information, FairExpand aims to scale fairness enforcement across the entire graph while preserving model utility.

\subsection{Phase 1: Partial Fairness Enforcement}\label{sec:p1} 

We begin by presenting Phase 1 of FairExpand. The objective of this phase is to enforce the partial individual fairness of the node embedding representations learned by $\mathcal{M}$, based on the observed partial similarity matrix $S^{(0)}$. To achieve this, we employ the bias in Equation~\eqref{equationbias}, computed only on $S^{(0)}$, as a regularizer of the utility loss $\mathcal{L}_{\text{utility}}$ that $\mathcal{M}$ would optimize in the absence of fairness considerations. Formally, we train $\mathcal{M}$ to minimize:
\begin{equation}\label{eq:L_NC}
    \mathcal{L}(\hat{Y}, S^{(0)}) = \mathcal{L}_{\text{utility}}(\hat{Y}) + \lambda \mathcal{L}_{\text{fairness}}(\hat{Y},S^{(0)}), 
\end{equation}
%akes as input the adjacency matrix $A$, the feature matrix $X$, and the partial similarity matrix $S^{(0)}$,
where the hyperparameter $\lambda > 0 $ balances fairness and utility, and:
\begin{equation}
    \mathcal{L}_{\text{fairness}}(\hat{Y},S^{(0)}) = \sum_{i=1}^n \sum_{j=1}^n \|\hat{y}_i - \hat{y}_j\|_{2}^2 S^{(0)}_{ij} = Tr(\hat{Y}^TL_{S^{(0)}}\hat{Y}).
\end{equation}

For example, in the case where $\mathcal{M}$ is a GNN trained for node classification, $\mathcal{L}(\hat{Y}, S^{(0)})$ would be minimized using gradient descent techniques \cite{goodfellow2016deep}, and $\mathcal{L}_{\text{utility}}$ would correspond to a classification loss comparing the predicted class probabilities in $\hat{Y}$ with the ground-truth labels in $Y$, such as the cross-entropy loss:
% \vspace{-1mm}
\begin{equation}
% \vspace{-1mm}
    \mathcal{L}_{\text{utility}}(\hat{Y}, Y) = - \sum_{i = 1}^n \sum_{c = 1}^{d_2} y_{ic}\log(\hat{y}_{ic}).
    \label{crossentropy}
    % \vspace{-1mm}
\end{equation}

\subsection{Phase 2: Similarity Information Expansion}\label{sec:p2} 

Although similarity information is only partially available, our objective is to enforce individual fairness across the entire graph. To address this challenge, Phase 2 of FairExpand aims to expand the observed similarity information to larger parts of the graph.

\subsubsection{Modeling Similarity Propagation as Link Prediction}

We model similarity information propagation as a link prediction task~\cite{kumar2020link}. As illustrated in Figure~\ref{fig:framework}, the partial similarity matrix $S^{(0)}$ naturally defines a sparse auxiliary similarity graph, $\mathcal{G}_{S^{(0)}}$, whose edges correspond to observed similarities between node pairs. Phase 2 of FairExpand consists in training an auxiliary link prediction model to infer additional similarity edges for this similarity graph, which we denote by $\mathcal{G}_{S^{(0)}}$. The partially fair representations $\hat{Y}$ obtained in Phase 1 serve as input features for this model, ensuring the individual fairness constraints enforced through Equation~\eqref{eq:L_NC} guide how new similarity edges are predicted.

However, two challenges arise when training this link predictor:
\begin{enumerate}
\item First, since a majority of the nodes in $\mathcal{G}_{S^{(0)}}$ are disconnected, it is impossible to distinguish between pairs of nodes that are truly dissimilar and pairs whose similarity is simply unobserved. As a result, our link prediction task is not a standard binary classification problem, but instead falls under the setting of positive-unlabeled (PU) learning~\cite{jaskie2019positive}.
\item Second, the sparsity of $\mathcal{G}_{S^{(0)}}$ turns this link prediction task into a few-shot learning problem, a setting in which many link predictors are known to struggle~\cite{bose2019meta}.
\end{enumerate}
 Without addressing these issues, the predictor would risk overfitting and fail to extract meaningful similarity information.

\subsubsection{PU-Learning-Based Link Prediction (PULL)}

To address the first challenge, we follow the methodology of PULL~\cite{kim_accurate_2024}, a link prediction method designed for positive–unlabeled graph settings. PULL behaves as if the  graph were fully connected: instead of considering only observed edges, it assigns a latent edge probability to every node pair, thereby treating observed edges as positive examples and other pairs as unlabeled rather than confirmed negatives.

An auxiliary graph convolutional network (GCN), trained to reconstruct edges while using  $\hat{Y}$ as node features, estimates edge probabilities on this graph. To avoid computing probabilities over all node pairs, using factorization strategies that allows edge probabilities to be estimated efficiently~\cite{kim_accurate_2024}. PULL then incrementally adds the top-$m_{\text{add}}$ most probable missing edges to the graph, for some  $m_{\text{add}} > 0$. Appendix \ref{app:pull} provides more technical details~about~PULL.

\subsubsection{$\epsilon$-Greedy PULL for Few-Shot Link Prediction}\label{egreedy}

Directly applying PULL is not suitable when $\mathcal{G}_{S^{(0)}}$ is highly sparse. In such cases, PULL tends to predict links almost exclusively between nodes that already participate in at least one observed similarity edge, while assigning an undifferentiated probability of $0.5$ to all remaining node pairs.
To address this second challenge and \textit{explore} larger regions of the graph, we propose \emph{$\epsilon$-Greedy PULL}, an extension of PULL which incorporates the $\epsilon$-Greedy strategy~\cite{sutton2018reinforcement} into the edge selection protocol. For some $\epsilon \in [0, 1]$, we add to $\mathcal{G}_{S^{(0)}}$:
\begin{enumerate}
\item The $(1-\epsilon) \cdot m_{\text{add}}$ highest-probability edges under PULL,
\item $\epsilon \cdot m_{\text{add}}$ randomly selected edges.
\end{enumerate}
This produces an expanded graph $\mathcal{G}_{S^{(1)}}$, corresponding to the updated partial similarity matrix $S^{(1)}$. Section \ref{math} and our experiments showcase and discuss the relevance of this exploration strategy.

\subsection{Iterative Alternation of Phase 1 and Phase 2}

\label{sec:p3}

As shown in Figure~\ref{fig:framework}, FairExpand alternates $K \geq 1$ times between Phase 1 and 2, that is, between refining  node representations based on the current similarity information and gradually expanding this information. 
At each step $k \in \{1, \dots, K\}$:
\begin{enumerate}
\item \textbf{Phase 1}: We produce the updated representations $\hat{Y}^{(k)}$ by enforcing partial individual fairness on the backbone model $\mathcal{M}$ based on the observed partial similarity matrix $S^{(k-1)}$.

\item \textbf{Phase 2}: Using $\epsilon$-Greedy PULL and based on $\hat{Y}^{(k)}$, we expand $S^{(k-1)}$ by adding $m_{\text{add}}$ edges, thereby obtaining the updated partial similarity matrix~$S^{(k)}$.
\end{enumerate}

The final vectors from $\hat{Y}^{(K)}$ are returned as the individually fair node embedding representations produced by FairExpand.

\noindent \textit{Remarks.} We note that, while a single iteration would suffice to produce representations, it would not propagate similarity information across the graph. Moreover, injecting a large number of pseudo-edges in one shot tends to degrade embedding quality in practice. These considerations motivate our iterative procedure in this work, which gradually refines $\hat{Y}^{(k)}$ and $S^{(k)}$  in a controlled~manner.

We also emphasize that, while the backbone GNN model $\mathcal{M}$ must be trained from scratch at $k=1$ using $A$, $X$, and $S^{(0)}$, for $k > 1$ it only needs to be fine-tuned. In these subsequent iterations, $\mathcal{M}$ is initialized with weights from the previous step, and node features $X$ are replaced with the current $\hat{Y}^{(k)}$. Finally, we acknowledge that one might also consider a heuristic-based stopping criterion instead of a fixed number of iterations $K$, which we leave for future work.

\subsection{Mathematical Intuition behind FairExpand}
\label{math}

To explain the effectiveness of FairExpand, we relate our $\varepsilon$-Greedy PULL exploration mechanism to recent work on fairness through representation smoothing~\cite{kumar_2021_center, peychev_2022_latent}. This connection highlights the importance of controlled noise for propagating fairness constraints.

\subsubsection{Graph-based Center Smoothing}
Center smoothing~\cite{kumar_2021_center} provides robustness guarantees for structured-output models by sampling perturbations $x+\delta$ of input point $x$, with $\delta\sim\mathcal{N}(0,\sigma^2 I)$, passing them through the model, and returning the center of the smallest ball containing at least half of the outputs. Peychev et~al.~\cite{peychev_2022_latent} adapt this idea to fairness over image data. Instead of injecting  noise, they generate similar individuals by interpolating along sensitive latent directions of a generative model and train their model, LASSI, so that the resulting representations collapse into a small $\ell_2$ ball. Our formulation generalizes the geometric intuition of center smoothing and LASSI to graphs but differs in several ways. First, rather than perturbing inputs, we construct neighborhoods from $S^{(0)}$ and its expanded variants~$S^{(k)}$, giving nodes a  evolving neighborhood that serves as a graph analogue of latent-space perturbations~\cite{peychev_2022_latent}. Moreover, unlike LASSI, our framework does not smooth classifier outputs but imposes a Lipschitz constraint directly on the embedding function, allowing any downstream classifier to inherit the induced smoothness.

\subsubsection{Local-Mean Interpretation of FairExpand.}
Formally, let us assume a binary masked similarity matrix of partial information. For notational brevity, we use $S = S^{(k)}$ and omit the superscript throughout this section. We can define the degree of node $i$ as $d_i = \sum_j S_{ij}$, its neighborhood as $N(i)=\{j : S_{ij}=1\}$, and let $E_S = \{(i,j): S_{ij}=1\}$. We show how our method can be viewed as a graph-based analog of LASSI, extending the geometric intuition of center smoothing to node embeddings. In this setting, the input $x$ corresponds to node features. Similarly, the Gaussian noise $(x + \delta)$ in the center smoothing approach is introduced via the $\varepsilon$ parameter of the $\varepsilon$-Greedy PULL fairness expansion protocol. Finally, the key element of center smoothing is the calculation of an enclosing ball. We will now introduce the notion of a \textit{localized mean} over a neighborhood $N(i) \in S^{(k)}$ and show how it is a graph-analogous formulation of the enclosing ball. First, since $S$ contains only binary elements defined by partial information edges, we can express the InFoRM bias as a summation over the partial information edges, $E_S$. Expanding the quadratic term and using the symmetry of $S$,  $\mathrm{Bias}(\hat Y, S)
= \sum_{i,j=1}^n \|\hat y_i - \hat y_j \|_2^2 S_{ij}
= \sum_{(i,j)\in E_S} \|\hat y_i - \hat y_j\|_2^2$
becomes:
\begin{equation}
\begin{aligned}
\mathrm{Bias}(\hat Y, S)
&= \sum_{(i,j)\in E_S}
\left(\|\hat y_i\|_2^2 + \|\hat y_j\|_2^2 - 2 \hat y_i^\top \hat y_j \right) \\
&= 2 \sum_{i} d_i \|\hat y_i\|_2^2
- 2 \sum_{i} \hat y_i^\top \left( \sum_{j\in N(i)} \hat y_j \right).
\end{aligned}
\end{equation}
By defining the localized mean
$\bar y_i = \frac{1}{d_i} \sum_{j\in N(i)} \hat y_j$, we~obtain:
\begin{equation}
\mathrm{Bias}(\hat Y, S)
= 2 \sum_{i} d_i \|\hat y_i\|_2^2
- 2 \sum_i d_i \,\hat y_i^\top \bar y_i,
\end{equation}
revealing that our optimization procedure encourages embeddings to contract around localized centers, mirroring how center smoothing collapses perturbed samples toward a robust center.

\subsection{Adaptability and Flexibility of FairExpand}

For clarity of exposition, we presented FairExpand in the specific setting where the model $\mathcal{M}$ is a GNN trained for node classification. Our experiments focus on this case as well, mainly using a GCN. However, we emphasize that the framework is  flexible. FairExpand can operate, not only with any other GNN architecture such as GraphSAGE or GAT \cite{hamilton2017representation,velivckovic2019graph}, but also with non-GNN graph representation learning models, as long as they output vectors $y_i$ \cite{hamilton2020graph}. This flexibility  implies that FairExpand does not impose a fixed computational complexity: its efficiency depends on the choice~of~$\mathcal{M}$.

Moreover, FairExpand is not limited to node classification. While $\mathcal{L}_{\text{utility}}$ was instantiated as a cross-entropy classification loss in Equation~\eqref{crossentropy}, it may be replaced with any other suitable objective, for example an edge reconstruction loss to optimize $y_i$ vectors for link prediction \cite{kumar2020link}. This adaptability allows FairExpand to be applied across a wide range of graph learning tasks and applications.

\section{Experimental Analysis} \label{sec:experiments}
This section presents our experimental evaluation of FairExpand. Our experiments are organized to answer the following questions:
\begin{itemize}[noitemsep,topsep=0pt] 
\item [\textbf{Q1.}] How effective is FairExpand in balancing node classification performance and individual bias mitigation? 
\item [\textbf{Q2.}] How do the key components of FairExpand contribute to its overall effectiveness? 

\end{itemize} 

\subsection{Experimental Settings}

\subsubsection{Datasets}
We conduct our experiments on six real-world, publicly available graphs of varying natures, sizes, and characteristics: the Coauthor, Pubmed, Cora, and Citeseer citation networks~\cite{schur_pitfalls_2018,sen2008collective, citeseer}, the Amazon purchase interaction graph~\cite{leskovec_dynamics_2007}, and the Flickr social network~\cite{zeng_graphSAINT-2020}. Table~\ref{tab:dataset} summarizes their key statistics, including the numbers of nodes, edges, classes, and features.

\subsubsection{Baselines.}
We compare FairExpand with a vanilla 2-layer GCN without fairness considerations and four individual fairness methods. These include the three graph-based frameworks InFoRM \cite{kang_inform_2020}, REDRESS \cite{dong_individual_2021}, and GFairHint \cite{xu_gfairhint_2023}, all designed for full similarity settings, as well as the non-graph method PFR~\cite{lahoti2019operationalizing}, which relaxes this assumption. PFR maximizes a trace objective using features $X$ and the partial similarity $S^{(0)}$. For FairExpand and  graph-based baselines (InFoRM, REDRESS, GFairHint), we set the backbone GNN to be the same as the vanilla model, i.e., a $2$-layer GCN.

\subsubsection{Implementation Details and Reproducibility.}
Due to space limitations, we report implementation details and full hyperparameter settings for FairExpand and all baselines in Appendix~\ref{ap:hp}. All experiments are conducted on a Linux server using one NVIDIA L40S GPU. The code, data, and requirements for reproducibility are open sourced at: \url{https://anonymous.4open.science/r/FairExpand-WWW-47BC/README.md}. 

\begin{table*}[t]
\centering
\caption{Node classification using FairExpand ($K=15$) and baselines. Scores are computed on test sets. Balance scores use the GCN without fairness considerations as the reference. Best balance scores are reported in bold and second best are underlined.}
% \vspace{-3mm}
\label{table:overall_performance}
\resizebox{0.975\linewidth}{!}{
\begin{tabular}{c|| cc|c ||cc|c ||cc|c }
\toprule 

\multirow{2}{*}{\textbf{Model}} & 
\multicolumn{3}{c}{\textbf{Coauthor}} & 
\multicolumn{3}{c}{\textbf{Amazon}} &
\multicolumn{3}{c}{\textbf{Flickr}} \\ 
\cmidrule(lr){2-4}\cmidrule(lr){5-7}\cmidrule(lr){8-10}
&F1 ($\uparrow$)& Bias ($\downarrow$) & \textbf{Balance} ($\uparrow$) & 
 F1 ($\uparrow$)& Bias ($\downarrow$) & \textbf{Balance} ($\uparrow$)& 
 F1 ($\uparrow$)& Bias ($\downarrow$) & \textbf{Balance} ($\uparrow$)\\ 
\midrule

GCN (Reference) & 
0.92 $\pm$ 0.01 & 32623.71 $\pm$ 3966.99 & --- &
0.93 $\pm$ 0.01 & 62277.01 $\pm$ 10965.37 & --- &
0.51 $\pm$ 0.01 & 10505.58 $\pm$ 3161.39 & --- \\ 

\midrule
PFR & 
0.46 $\pm$ 0.06 & 20.87 $\pm$ 3.68 & 0.64 $\pm$ 0.04 &
0.39 $\pm$ 0.02 & 33.44 $\pm$ 2.11 & 0.60 $\pm$ 0.01 &
{0.44 $\pm$ 0.00} & {26.23 $\pm$ 5.29} & \textbf{0.89 $\pm$ 0.01} \\ 

InFoRM & 
{0.88 $\pm$ 0.03} & {8021.25 $\pm$ 1619.40} & \underline{0.81 $\pm$ 0.03} &
{0.85 $\pm$ 0.02} & {8981.56 $\pm$ 2125.72} & \underline{0.69 $\pm$ 0.04} &
0.38 $\pm$ 0.14 & 905.64 $\pm$ 1326.24 & 0.79 $\pm$ 0.24 \\

REDRESS & 
0.32 $\pm$ 0.09 & 23953.98 $\pm$ 21232.26 & 0.21 $\pm$ 0.14 &
0.24 $\pm$ 0.02 & 22013.31 $\pm$ 15573.05 & 0.07 $\pm$ 0.03 &
0.43 $\pm$ 0.02 & 1800746.47 $\pm$ 0.00 & 0.59 $\pm$ 0.03 \\

GFairHint & 
0.91 $\pm$ 0.00 & 657745.01 $\pm$ 399282.36 & 0.69 $\pm$ 0.00&
0.66 $\pm$ 0.15 & 956512.86 $\pm$ 938945.62 & 0.50 $\pm$ 0.12 &
0.27 $\pm$ 0.18 & 33579.90 $\pm$ 55016.65 & 0.46 $\pm$ 0.30 \\

\midrule
FairExpand & 
0.78 $\pm$ 0.05 & 2779.55 $\pm$ 692.52 & \textbf{0.84 $\pm$ 0.06} &
0.81 $\pm$ 0.08 & 5712.18 $\pm$ 1441.22 & \textbf{0.73 $\pm$ 0.07} &
0.42 $\pm$ 0.00 & 40.66 $\pm$ 25.80 & \underline{0.88 $\pm$ 0.00} \\

\midrule
\midrule
\multirow{2}{*}{\textbf{Model}} &
\multicolumn{3}{c}{\textbf{Pubmed}} & 
\multicolumn{3}{c}{\textbf{Cora}} &
\multicolumn{3}{c}{\textbf{Citeseer}} \\ 
\cmidrule(lr){2-4}\cmidrule(lr){5-7}\cmidrule(lr){8-10}
&F1 ($\uparrow$)& Bias ($\downarrow$) & \textbf{Balance} ($\uparrow$) &
 F1 ($\uparrow$)& Bias ($\downarrow$) & \textbf{Balance} ($\uparrow$) &
 F1 ($\uparrow$)& Bias ($\downarrow$) & \textbf{Balance} ($\uparrow$)\\ 
\midrule

GCN (Reference) & 
0.87 $\pm$ 0.01 & 2093.56 $\pm$ 54.17 & --- &
0.86 $\pm$ 0.01 & 359.65 $\pm$ 45.13 & --- &
0.76 $\pm$ 0.01 & 111.46 $\pm$ 22.59 & ---\\ 

\midrule
PFR & 
0.55 $\pm$ 0.03 & 1.73 $\pm$ 0.95 & \underline{0.74 $\pm$ 0.03} &
0.37 $\pm$ 0.00 & 0.00 $\pm$ 0.00 & 0.60 $\pm$ 0.00 &
0.24 $\pm$ 0.03 & 0.00 $\pm$ 0.00 & 0.51 $\pm$ 0.03 \\

InFoRM & 
0.86 $\pm$ 0.01 & 1526.11 $\pm$ 151.93 & 0.71 $\pm$ 0.03 &
{0.85 $\pm$ 0.01} & {238.85 $\pm$ 40.40} & \underline{0.72 $\pm$ 0.05} &
{0.75 $\pm$ 0.01} & {126.05 $\pm$ 17.03} & \underline{0.70 $\pm$ 0.01} \\

REDRESS & 
0.44 $\pm$ 0.02 & 2.04 $\pm$ 1.89 & 0.57 $\pm$ 0.03 &
0.37 $\pm$ 0.08 & 0.24 $\pm$ 0.15 & 0.53 $\pm$ 0.11 &
0.47 $\pm$ 0.03 & 5.65 $\pm$ 1.46 & 0.64 $\pm$ 0.12 \\

GFairHint & 
0.85 $\pm$ 0.00 & 39525.90 $\pm$ 12604.95 & 0.69 $\pm$ 0.00 &
0.82 $\pm$ 0.01 & 5412.02 $\pm$ 1315.65 & 0.67 $\pm$ 0.01 &
0.69 $\pm$ 0.01 & 4219.85 $\pm$ 1652.74 & 0.64 $\pm$ 0.01 \\

\midrule
FairExpand & 
0.86 $\pm$ 0.00 & 1425.01 $\pm$ 63.92 & \textbf{0.75 $\pm$ 0.03} & 
0.85 $\pm$ 0.02 & 261.01 $\pm$ 48.85 & \textbf{0.73 $\pm$ 0.01} &
0.76 $\pm$ 0.00 & 76.92 $\pm$ 0.00 & \textbf{0.78 $\pm$ 0.01} \\

\bottomrule
\end{tabular}
}
% \vspace{-5mm}
\end{table*}

\subsubsection{Evaluation.} We consider node classification as the downstream evaluation task. All graph datasets include ground-truth node labels (ranging from 3 to 15 classes) used for evaluation. We generate train/validation/test splits for each dataset using a 60\%/20\%/20\% ratio. To evaluate utility, we compute the multi-class F1 score on test nodes~\cite{manning2009introduction}. To evaluate individual fairness, we compute $\text{Bias}(\hat{Y}, S_{\rm test})$ (as defined in Equation~\eqref{equationbias}), where $S_{\rm test}$ is a masked version of the full similarity matrix $S$ containing pairwise values for only test nodes. To assess the \textit{trade-off between fairness and utility}, we introduce a balance score defined as a weighted average of the relative F1 score preserved and the relative bias mitigated with respect to the backbone GCN model without fairness considerations, used as a reference:
  
\begin{equation}\label{eq:balance}
    \text{Balance} =  \biggl[ \alpha \cdot \underbrace{\left(\frac{\text{F1}_\text{Model}}{\text{F1}_{\text{Ref}}}\right)}_{\text{\% F1 preserved}} + (1-\alpha) \cdot\underbrace{\left(\frac{\text{Bias}_{\text{Ref}} - \text{Bias}_{\text{Model}}}{\text{Bias}_{\text{Ref}}} \right)}_{\text{\% bias mitigated}} \biggr],
       %\vspace{-.5em}
\end{equation}

where $\alpha \in [0,1]$ balances the importance of utility and fairness. A model achieving a good trade-off should preserve the utility of the reference model while reducing its bias as much as possible. In practice, we cap each percentage of Equation \eqref{eq:balance} to stay within $[0,1]$, ensuring that the balance score also remains in this range. Under this scheme, a perfect balance score of 1 indicates that a model completely mitigates the reference GCN's bias without any performance degradation. We set $\alpha = 0.7$ in our experiments.

Finally, to analyze the effect of our $\epsilon$-Greedy exploration mechanism, we introduce the \emph{Node Overlap Ratio (NOR)}, which measures the fraction of nodes in the updated partial similarity graph that were already present in the initial graph:
\begin{equation}\label{eq:NOR}
    \text{Node Overlap Ratio (NOR)} = \frac{|\mathcal{V}^{(0)} \cap \mathcal{V}^{(K)}|}{|\mathcal{V}^{(K)}|},
\end{equation}
where $\mathcal{V}^{(k)}$ denotes the set of nodes incident to edges in the partial similarity graph $\mathcal{G}_{S^{(k)}}$ at iteration $k$. A lower NOR indicates that similarity propagation reaches a broader portion of the graph because more nodes were introduced between iterations $0$ and $K$.

\subsubsection{Similarity Graph Construction.} \label{sec:building_S}

To simulate the ground-truth similarity matrix $S$ in our experiments, we follow standard practices used in prior studies on individual fairness for graphs~\cite{kang_inform_2020, xu_gfairhint_2023, dong_individual_2021}. We compute the pairwise cosine similarity of node features $X$, i.e., $S_{ij} = \frac{x^\top_i x_j}{\|x_i\| \|x_j\|}$, for all $(i,j) \in \mathcal{V}^2$. Then, to obtain the partial similarity matrix $S^{(0)}$ of sparse expert annotations, we filter out similarity values below a threshold $\tau \in (0,1)$ and randomly sample a subset of the remaining pairs. A more detailed discussion of this procedure, including full pseudocode, is provided in Appendix~\ref{sec:impl}.

\noindent \textit{Remark.}
In our experiments, we generate a different $S^{(0)}$ graph for each seed. This is purposefully done to showcase the stability of our method to the stochasticity of an expert annotation procedure, since otherwise we cannot guarantee that the observed performance gains are not simply due to a particular choice of $S^{(0)}$.

\begin{figure}[h!]
    \centering
    \includegraphics[trim= 80 0 50 0,clip,width=.8\linewidth]{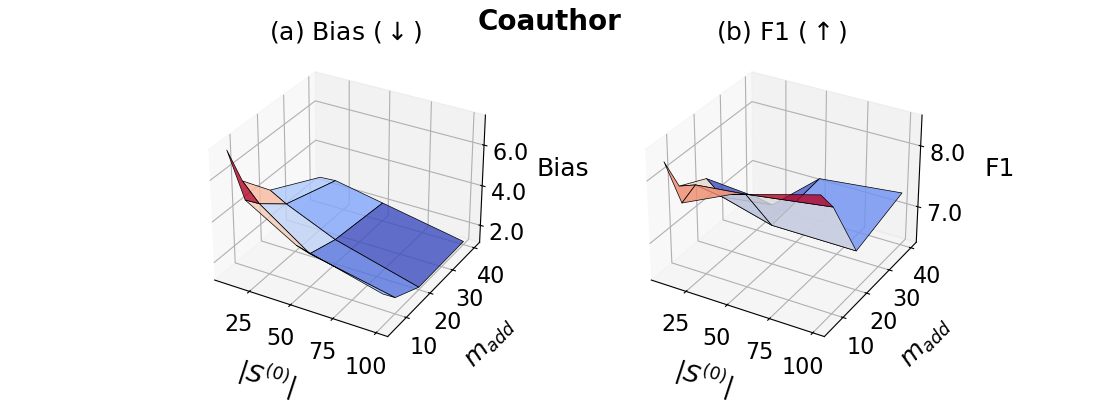}
    \includegraphics[trim= 80 0 50 0,clip,width=.8\linewidth]{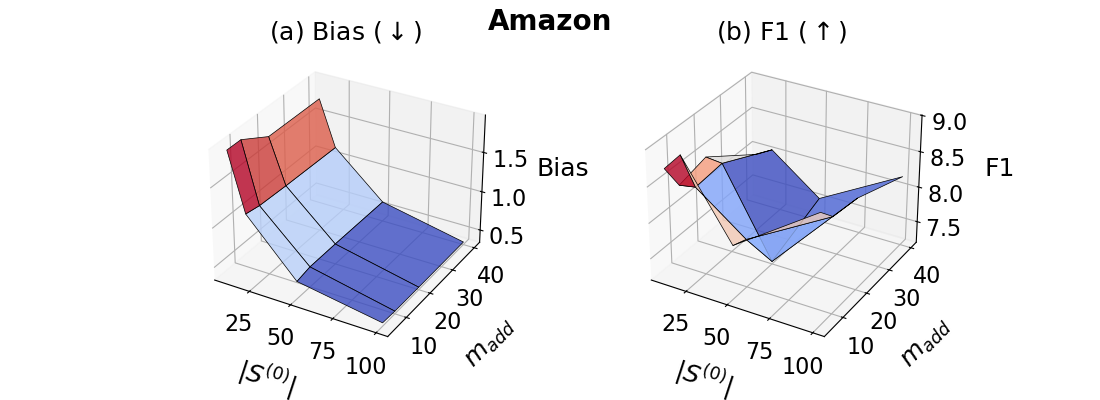}
    \includegraphics[trim= 80 0 50 0,clip,width=.8\linewidth]{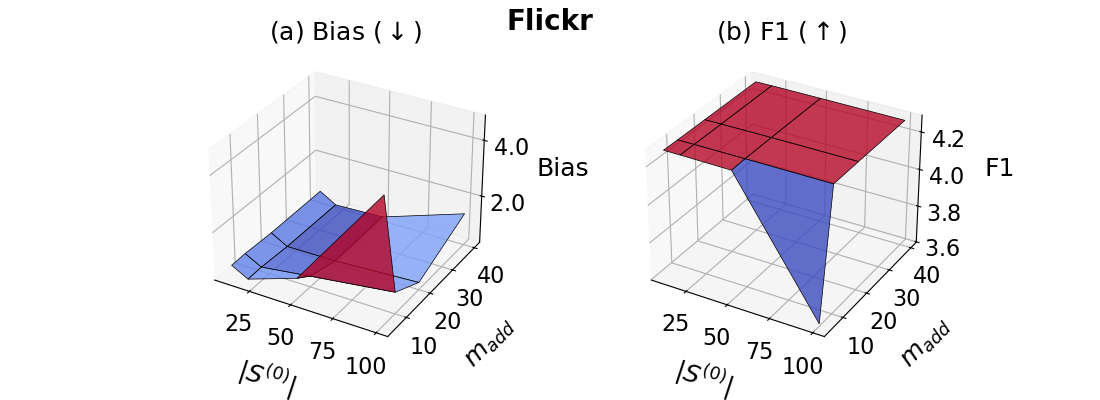}
    \includegraphics[trim= 80 0 50 0,clip,width=.8\linewidth]{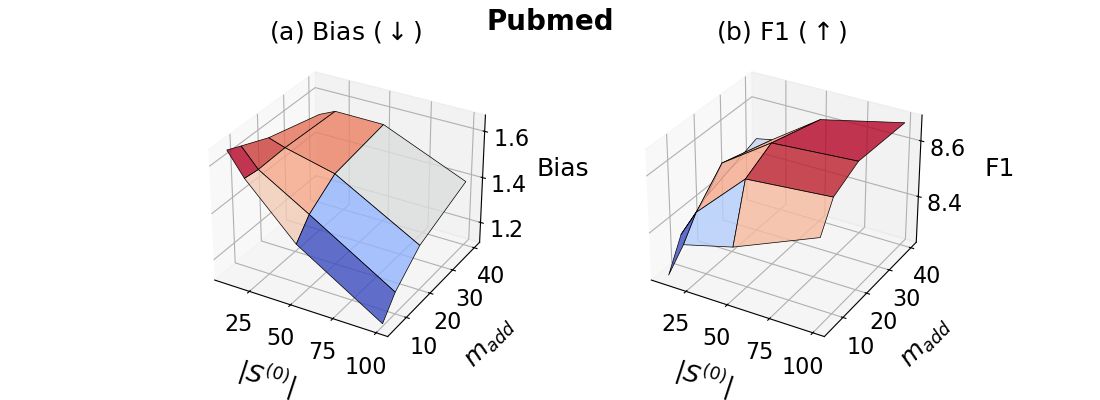}
    \includegraphics[trim= 80 0 50 0,clip,width=.8\linewidth]{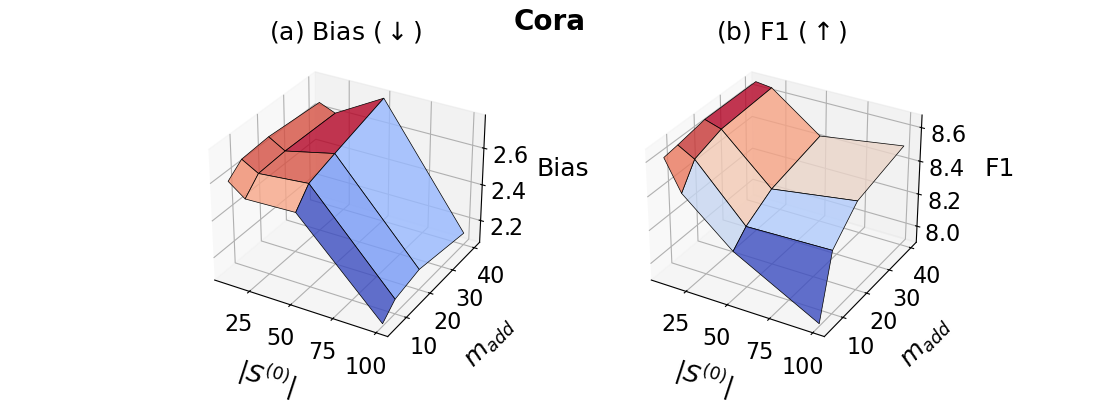}
    \includegraphics[trim= 80 0 50 0,clip,width=.8\linewidth]{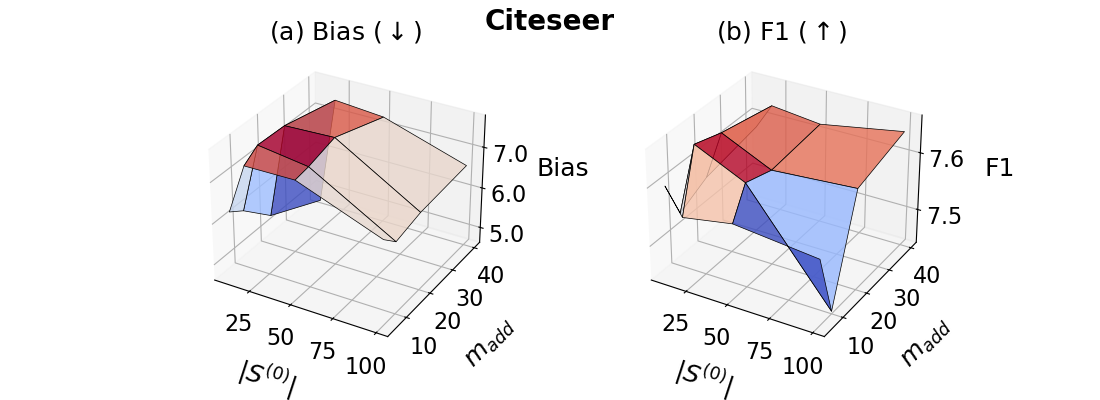}
    % \vspace{-1em}
    \caption{Sensitivity analysis of F1 and bias scores with respect to the number of similarities in $S^{(0)}$, denoted $|S^{(0)}|$, and the number of edges $m_{\text{add}}$ added in Phase 2 of FairExpand.}
    \label{fig:sensitivity}
    % \vspace{-2em}
\end{figure}

\subsection{Q1: Balancing Performance and Fairness}

Table~\ref{table:overall_performance} reports the results of all methods across the six datasets. We primarily focus on the balance metric, which reflects each method’s ability to preserve the node classification performance of the reference GCN without fairness enforcement while reducing its individual fairness bias. Overall, FairExpand achieves the highest balance score on five out of six graphs and the second highest on Flickr, within the margin of error of the best method.

\begin{figure}[]
\centering

% ---- TOP ROW ----
\begin{subfigure}{\linewidth}
    \centering
    \includegraphics[width=.7\linewidth, trim={2cm 3cm 9cm 0cm},clip]{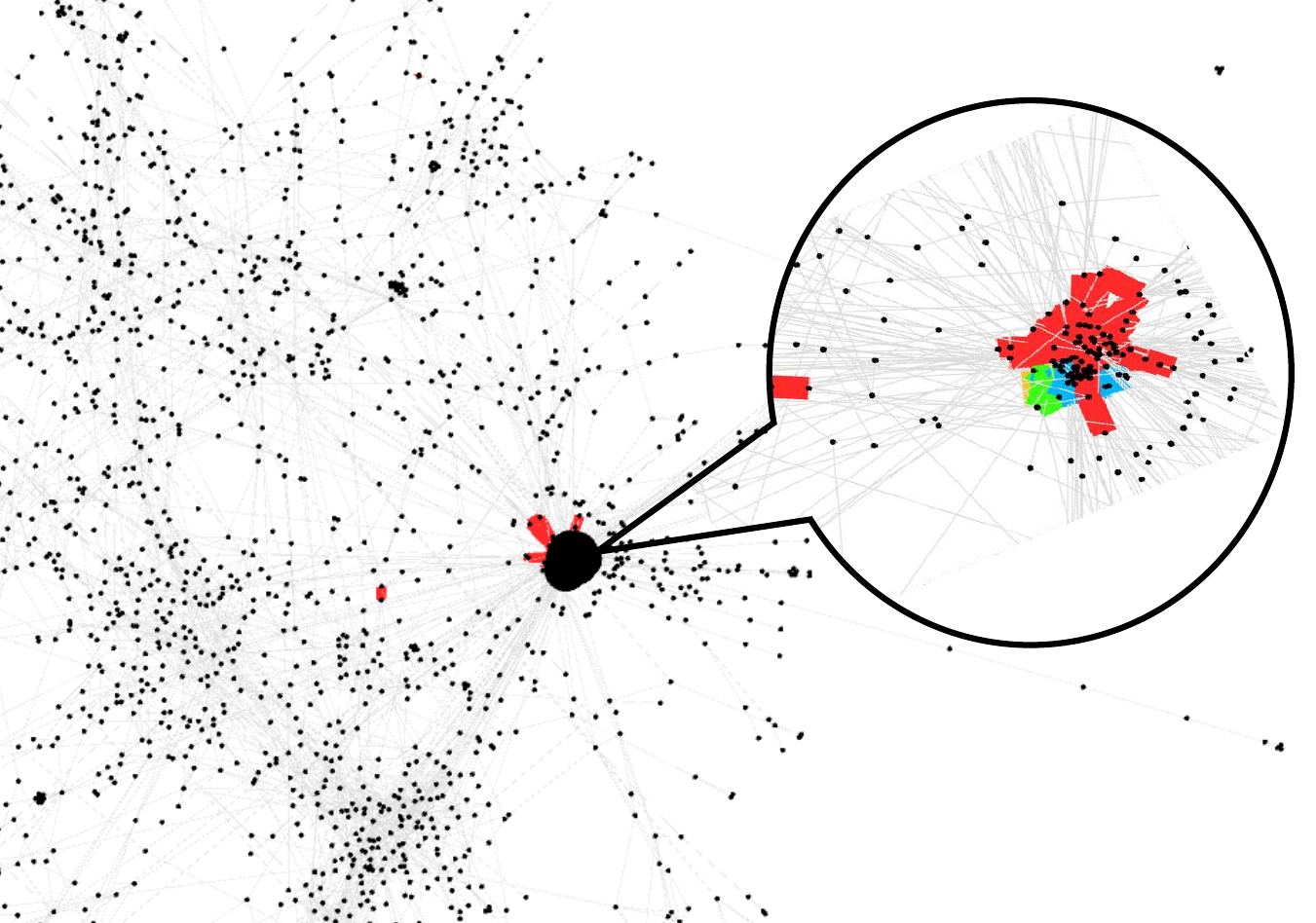}
    \caption{Citeseer}
    \includegraphics[width=.9\linewidth]{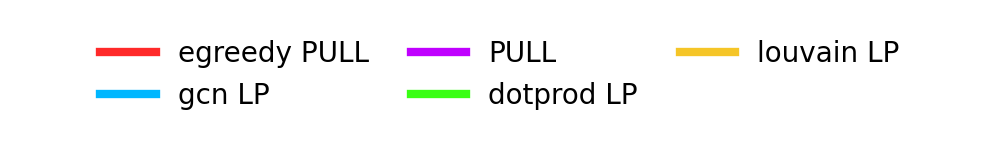}
\end{subfigure}

% ---- BOTTOM ROW ----
\begin{subfigure}{0.7\linewidth}
    \centering
    \includegraphics[width=\linewidth]{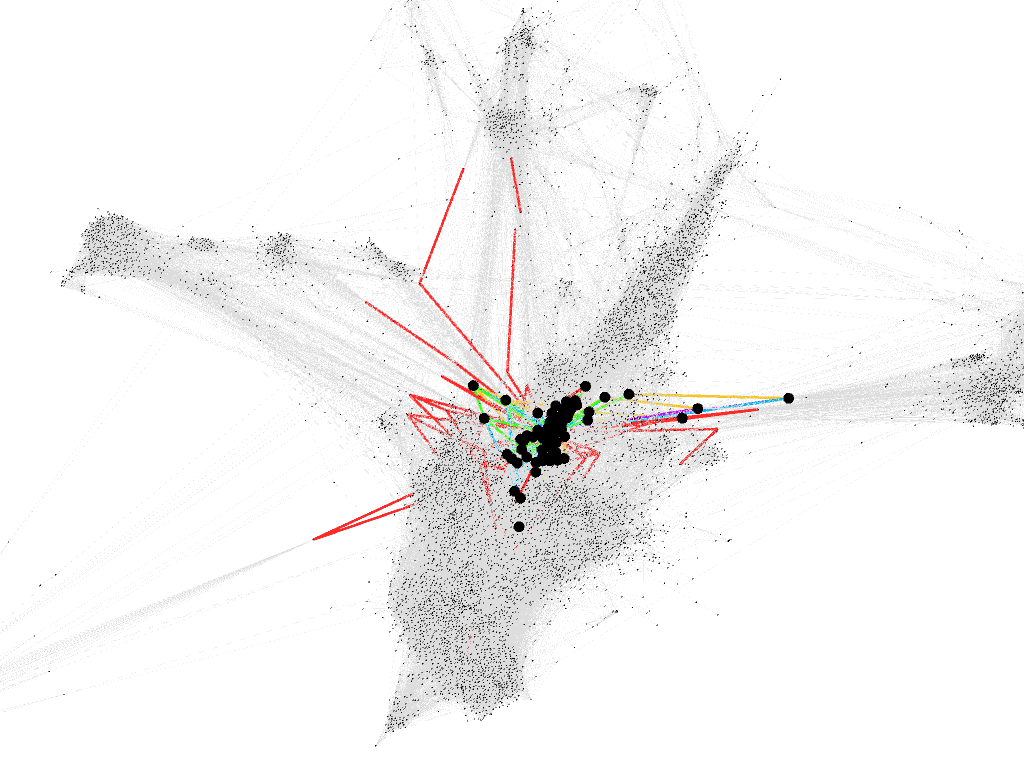}
    \caption{Amazon}
\end{subfigure}
\caption{Exploration trajectories after a Phase 2 iteration in FairExpand ablations. Larger nodes indicate original members of $S^{(0)}$. We observe that $\varepsilon$-greedy PULL (red edges) exhibits the broadest exploration patterns.}
\label{fig:graph_viz}

% \vspace{-2.3em}
\end{figure}
A key factor behind this success is FairExpand’s capacity to expand similarity information to regions of the graph lacking initial fairness annotations. Compared with InFoRM, which would correspond to FairExpand's Phase~1 only, we observe that FairExpand substantially reduces bias with minimal performance loss. This confirms that the combination of similarity propagation and representation refinement (Phase~2) significantly enhances individual fairness, even when similarity information is sparse.

When compared to PFR, which is designed for partial information settings but ignores the graph structure, the advantage of FairExpand becomes even clearer. PFR attains low bias only at the cost of a severe drop in utility, whereas FairExpand maintains strong performance while achieving meaningful fairness improvements.

REDRESS, by contrast, exhibits instability in the partial information setting. This stems from its reliance on \textit{relative} pairwise distances, which enforce ranking consistency even when only a small subset of true similarity relationships is known. In settings where only the edges in $S^{(0)}$ represent similarity, this leads to unreliable fairness corrections. GFairHint shows similar issues: the sparsity of observed similarity information degrades the quality of its fairness hints, resulting in consistently low performance. Overall, these comparisons illustrate that FairExpand uniquely leverages both similarity structure and graph connectivity, enabling fairness improvements across various settings while jointly preserving performance, even when fairness annotations are incomplete.

\begin{table*} [h]
\centering
\caption{Ablation study of FairExpand components for node classification. The acronyms ``P1'' and ``P2'' denote Phase 1 and Phase 2 of FairExpand, respectively. Lower NOR scores indicate broader similarity propagation across nodes during Phase 2. }%Bold indicates model that achieved best test bias.}
% \vspace{-1em}
% \vspace{-2mm}
\label{table:ablation}
% \vspace{-2mm}
\resizebox{0.975\linewidth}{!}{
\begin{tabular}{l || ccc || ccc || ccc}
\toprule 

\multirow{2}{*}{\textbf{FairExpand Ablations}} &
\multicolumn{3}{c}{\textbf{Coauthor}} & 
\multicolumn{3}{c}{\textbf{Amazon}} &
\multicolumn{3}{c}{\textbf{Flickr}} \\ 
\cmidrule(lr){2-4}\cmidrule(lr){5-7}\cmidrule(lr){8-10}
&F1 ($\uparrow$)& Bias ($\downarrow$) & NOR ($\downarrow$) &
F1 ($\uparrow$)& Bias ($\downarrow$) & NOR ($\downarrow$) &
F1 ($\uparrow$)& Bias ($\downarrow$) & NOR ($\downarrow$) \\ 
\midrule
GCN & 
0.92 $\pm$ 0.01 &  32623.71 $\pm$ 3966.99 & --- &
0.93 $\pm$ 0.01 & 62277.01 $\pm$ 10965.37 & --- &
0.51 $\pm$ 0.01 & 10505.58 $\pm$ 3161.39 & --- \\  

GCN + P1 & 
0.88 $\pm$ 0.03 & 8021.25 $\pm$ 1619.40 &--- &
0.85 $\pm$ 0.02 & 8981.56 $\pm$ 2125.72 & --- &
0.38 $\pm$ 0.14 & 905.64 $\pm$ 1326.24  & --- \\
{GCN + P1 + P2: Dot Product} & 
0.78 $\pm$ 0.06 & 5385.53 $\pm$ 1183.73  & 1.00 $\pm$ 0.00 &
0.81 $\pm$ 0.08 & 8453.17 $\pm$ 1800.75 & 1.00 $\pm$ 0.00 &
0.42 $\pm$ 0.00 & 211.4 $\pm$ 330.66 & 0.86 $\pm$ 0.14 \\ 
{GCN + P1 + P2: GCN} & 
0.75 $\pm$ 0.11 & 5942.10 $\pm$ 1392.58 & 1.00 $\pm$ 0.00 &
0.82 $\pm$ 0.10 & 8672.64 $\pm$ 1919.22 & 1.00 $\pm$ 0.00 &
0.43 $\pm$ 0.02 & 178.70 $\pm$ 277.49 & 1.00 $\pm$ 0.00 \\ 
GCN + P1 + P2: PULL & 
0.82 $\pm$ 0.02 & 5141.42 $\pm$ 1034.42 & 1.00 $\pm$ 0.00 &
0.83 $\pm$ 0.11 & 8467.44 $\pm$ 1745.54 & 1.00 $\pm$ 0.00 &
0.37 $\pm$ 0.14 & 858.41 $\pm$  1413.96& 1.00 $\pm$ 0.00 \\ 
GCN + P1 + P2: Louvain PULL & 
0.79 $\pm$ 0.05 & 4800.02 $\pm$ 778.23 & 1.00 $\pm$ 0.00 &
0.85 $\pm$ 0.08 & 8457.32 $\pm$ 1776.45 & 1.00 $\pm$ 0.00 &
0.42 $\pm$ 0.00& 67.36 $\pm$ 35.09 & 1.00 $\pm$ 0.00 \\
\midrule 
GCN + P1 + P2: $\epsilon$-Greedy PULL & 
{0.78 $\pm$ 0.05} & {2779.55 $\pm$ 692.52} & {0.57 $\pm$ 0.01} &
{0.81 $\pm$ 0.08} & {5712.18 $\pm$ 1441.22} & {0.58 $\pm$ 0.01} &
{0.42 $\pm$ 0.00} & {40.66 $\pm$ 25.80} & {0.40 $\pm$ 0.00} \\  
\midrule
\midrule

\multirow{2}{*}{\textbf{FairExpand Ablations}} &
\multicolumn{3}{c}{\textbf{Pubmed}} & 
\multicolumn{3}{c}{\textbf{Cora}} &
\multicolumn{3}{c}{\textbf{Citeseer}} \\ 
\cmidrule(lr){2-4}\cmidrule(lr){5-7}\cmidrule(lr){8-10}
&F1 ($\uparrow$)& Bias ($\downarrow$) & NOR ($\downarrow$) &
F1 ($\uparrow$)& Bias ($\downarrow$) & NOR ($\downarrow$) &
F1 ($\uparrow$)& Bias ($\downarrow$) & NOR ($\downarrow$) \\ 
\midrule
GCN & 
0.87 $\pm$ 0.01 & 2093.56 $\pm$ 54.17  & --- &
0.86 $\pm$ 0.01 & 359.65 $\pm$ 45.13& --- &
0.75 $\pm$ 0.01 & 106.46 $\pm$ 22.59 & --- \\  

GCN + P1 & 
0.87 $\pm$ 0.00 & 1942.50 $\pm$ 85.18 & --- &
0.86 $\pm$ 0.00 & 294.75 $\pm$ 53.83  & --- &
0.76 $\pm$ 0.01 & 106.05 $\pm$ 24.70 & --- \\

{GCN + P1 + P2: Dot Product} & 
0.79 $\pm$ 0.07 & 1609.91 $\pm$ 241.23 & 1.00 $\pm$ 0.00 &
0.81 $\pm$ 0.04& 253.24 $\pm$ 52.69 & 1.00 $\pm$ 0.00 &
0.74 $\pm$ 0.01 & 75.61 $\pm$ 17.59 & 1.00 $\pm$ 0.00 \\ 

{GCN + P1 + P2: GCN} & 
0.82 $\pm$ 0.02 & 1608.18 $\pm$ 88.50  & 0.98 $\pm$ 0.04  &
0.81 $\pm$ 0.02 & 252.72 $\pm$ 47.11& 1.00 $\pm$ 0.00&
0.75 $\pm$ 0.01 & 74.21 $\pm$ 21.77 &  1.00 $\pm$ 0.00 \\ 

GCN + P1 + P2: PULL & 
0.83 $\pm$ 0.06 & 1699.63 $\pm$ 58.62 & 1.00 $\pm$ 0.00 &
0.84 $\pm$ 0.01 & 265.31 $\pm$ 51.34  & 1.00 $\pm$ 0.00 &
0.75 $\pm$ 0.00 & 82.29 $\pm$ 20.22 & 1.00 $\pm$ 0.00 \\
GCN + P1 + P2: Louvain PULL & 
{0.83 $\pm$ 0.06} & {1665.63 $\pm$ 69.38} &  {1.00 $\pm$ 0.00} &
0.84 $\pm$ 0.03 & 263.14 $\pm$ 51.02 & 1.00 $\pm$ 0.00 &
0.75 $\pm$ 0.01 & 81.37 $\pm$ 20.75 & 1.00 $\pm$ 0.00 \\
\midrule 
GCN + P1 + P2: $\epsilon$-Greedy PULL & 
0.84 $\pm$ 0.03 & 1552.77 $\pm$ 103.94 & 0.39 $\pm$ 0.08 &
0.85 $\pm$ 0.02 & 261.01 $\pm$ 48.85 & 0.38 $\pm$ 0.00 &
0.76 $\pm$ 0.00 & 76.92 $\pm$ 0.00 & 0.39 $\pm$ 0.00 \\  
\bottomrule
\end{tabular}
}
% \vspace{-5mm}
\end{table*}

\subsection{Q2: Analysis of FairExpand} \label{sec:ablation}

\subsubsection{Ablation Study}
FairExpand consists of two key phases: Partial Fairness Enforcement (Phase 1) and Similarity Information Expansion (Phase 2). We perform in Table~\ref{table:ablation} an ablation study to evaluate the contribution of each phase to the overall effectiveness of the framework. For Phase 2, we also compare our proposed $\epsilon$-Greedy PULL link predictor used in FairExpand with four alternative link prediction variants.

From Table~\ref{table:ablation}, we make several observations. First, comparing GCN and GCN + Phase 1 (rows 1--2) shows that incorporating Phase 1  alone already reduces bias substantially across all datasets (e.g., from 32623.71 to 8021.25 on Coauthor). Second, adding Phase~2 and integrating it into the iterative expansion procedure (rows 3--6) yields further bias reduction across all graphs (e.g., down to 2779.55 on Coauthor). We compare four alternatives for link prediction in Phase 2: dot product, a GCN-based predictor, the original PULL implementation (without randomization), and a Louvain community-based method that prioritizes inter-community edge exploration \cite{blondel2008fast}.
Since the goal of Phase~2 is to \textit{expand} partial similarity information, we focus on the NOR metric as the primary indicator for differentiating link prediction strategies. Results show that $\varepsilon$-Greedy PULL, unlike the other predictors, introduces new nodes at $K$, so that the final similarity graph $S^{(K)}$ contains more than 40\% (i.e., $1-\text{NOR}$) newly exposed nodes on all datasets. This property is crucial in partial information settings: exposing a broader portion of the population to the fairness mechanism enables genuine expansion of similarity knowledge. Figure~\ref{fig:graph_viz} illustrates the substantial topological differences across several selected datasets. Due to space considerations, we showcase only Amazon and Citeseer as representative examples. We find that Coauthor, Flickr, and Pubmed resemble Amazon in exhibiting well-formed community structures while Cora and Citeseer are far less segmented, which makes meaningful exploration more challenging. We also present a further \textit{magnified} image of Citeseer to highlight the superior exploration patterns of our method. In both the visualized datasets, we can see that $\varepsilon$-Greedy explores a substantially broader portion of the overall graph in comparison with other link prediction methods. Additionally, Table~\ref{table:ablation} shows that $\epsilon$-Greedy PULL achieves the lowest bias on the four largest graphs (Coauthor, Amazon, Flickr, and Pubmed), while the GCN link predictor performs best on the two smaller datasets, Cora and Citeseer. Table~\ref{table:tau} and Figure \ref{fig:distributions} (see Appendix \ref{ap:distributions}) further indicate that Citeseer and Cora contain fewer highly similar node pairs on average, leading to a lower selected threshold $\tau$ in our experiments, and a more challenging exploration landscape. 

% \vspace{-1em}

\subsubsection{Sensitivity Analysis}
We can further contextualize the differences in dataset performance by examining the sensitivity analysis shown in Figure~\ref{fig:sensitivity}. These plots reveal that our method induces distinct relationships between F1 and bias across the various datasets. In the Amazon dataset, we observe a moderately positive association between bias and F1, suggesting that bias mitigation and node classification accuracy are not always in direct conflict. In contrast, the remaining datasets exhibit a clearly inverse relationship, indicating that stronger bias mitigation, achieved through larger values of $|S^{(0)}|$ and $m_{\text{add}}$, consistently reduces F1 performance.

Furthermore, comparing Flickr and PubMed, Coauthor, Cora, and Citeseer shows that in Flickr this trade-off occurs sharply at a specific parameter setting, implying that the dataset does not require extensive tuning. In PubMed or Cora, however, the trade-off landscape is smoother, offering more flexibility in controlling the balance between fairness and utility. Taken together, these results demonstrate that FairExpand provides a reliable and tunable mechanism for navigating fairness and utility trade-offs across fundamentally different graph structures.

\section{Conclusion}\label{conclusion}
This paper introduced FairExpand, a flexible framework for individual fairness on graphs and the first to enable effective learning of individually fair node embedding representations when only partial similarity information is available. FairExpand operates through a two-step iterative pipeline that alternates between refining embeddings using a backbone model and gradually expanding similarity information. In particular, we present the $\varepsilon-$Greedy PULL link predictor which is able to successfully interpolate between exploration (integrating novel individuals into the partial similarity information) and exploitation (leveraging known similarities to further reinforce individual fairness). As demonstrated through extensive experiments on six real-world graph datasets, this procedure consistently enhances individual fairness across the entire graph while preserving node classification performance. We note that FairExpand is a flexible framework which can be used with a wide range of backbone models on a wide array of tasks beyond those explored in this work. These results position FairExpand as a practical and scalable approach for enabling graph-based individual fairness in real-world settings with limited similarity annotations. 

Several methodological extensions are also possible. One direction is to introduce a principled stopping criterion for the iterative pipeline, instead of fixing $K=15$ as done in this paper. Another is to incorporate active learning, where the edges added at each iteration $k$ are selected through a parameterized neural mechanism rather than through predefined heuristics~\cite{xinyu_sensitivity_2024}.

\clearpage

\bibliographystyle{ACM-Reference-Format}
\bibliography{reference}

\clearpage
\section*{Appendix}
\appendix

\section{Implementation Details}\label{sec:impl}
\subsection{Details on PULL Link Prediction (Phase 2)}
\label{app:pull}
PULL's strength lies in its ability to extrapolate from a set of positive and \textit{unlabeled} edges, where the unlabeled set considers all possible node pairs in the graph. To enable this, each pair $(i,j)$ is associated with a latent variable $p(e_{ij})$, indicating the probability of an edge: 
\begin{equation}\label{eq:edge}
    p(e_{ij}) = \begin{cases} 1, & \text{if}~~~e_{ij} \in E_{S^{(0)}} \\ 
    PULL(\hat{y_i}) \cdot PULL(\hat{y_j})& \text{otherwise.}      
    \end{cases}
\end{equation}
We use an auxiliary 2-layer GCN as the PULL link predictor and treat the learned representations $\hat{Y}$ from Phase 1 as node features. At each training iteration, PULL selects edges with the highest $p(e_{ij})$  probabilities to integrate into the graph. To avoid computing probabilities over all node pairs, PULL uses Markov chain strategy that allows edge probabilities to be estimated efficiently~\cite{kim_accurate_2024}. Algorithm \ref{alg:PULL} shows how PULL is used in our setting.  

\begin{algorithm}[h]
\caption{Training of PULL  (without $\varepsilon$-greedy)}\label{alg:PULL}
\begin{algorithmic}[1]
\Procedure{Training Loop}{$S^{(0)}$, $\hat{Y}$, $m_{\text{add}}$}
\For{epoch = 1 to $T$}
    \If {epoch = 1}
        \State $z  \gets \mathrm{PULL.encode}(S^{(0)}, \hat{Y}$)
    \Else 
        \State new\_edges $\gets \mathrm{PULL}.\mathrm{calc\_prob}(z, S^{(0)}, m_{\text{add}})$
        \State $S^{(0)} \gets \mathrm{PULL}.\mathrm{merge\_edges}(S^{(0)}, \mathrm{new\_edges})$
    \EndIf
    \For{$inner = 1$ to $200$}
        \State $z \gets \mathrm{PULL.encode}(S^{(0)}, z)$
        \State neg\_edge $\gets \mathrm{negative\_sampling}(S^{(0)})$
        \State pos\_edge $\gets S^{(0)}.\text{edges}()$
        \State edge\_label = pos\_edge + neg\_edge
        \State edge\_pred $\gets \mathrm{PULL.decode}(z, \mathrm{edge\_label})$
        \State $\mathcal{L} \gets \mathrm{BCEWithLogits}( \mathrm{edge\_pred, edge\_label})$
        \State Backpropagate $\mathcal{L}$ and update weights
    \EndFor
\EndFor
\EndProcedure
\end{algorithmic}
\end{algorithm}

\subsection{Construction of $S^{(0)}$}
\label{appendixsimilarity}
Algorithm~\ref{alg:subpopsampling} details the  procedure used to generate $S^{(0)}$ from $S$. Line 2 computes pairwise cosine similarities over input features $X$. Line 3 filters them based on $\tau$ to simulate an expert's knowledge of similar individuals. Line 4 randomly selects edges from the pool of similar individuals. Line 5 constructs $S^{(0)}$ using these edges. 

\begin{algorithm}[h!]
\caption{Construction of $S^{(0)}$}\label{alg:subpopsampling}
\hspace*{-1.7cm} \textbf{Input}: $\mathcal{G} = (\mathcal{V},\mathcal{E})$, $X$, similarity threshold $\tau$ \\
\hspace*{-5.7cm}\textbf{Output}: $S^{(0)}$
\begin{algorithmic}[1]
\Procedure{Sparse $S$-Construction}{$\mathcal{G}, X, \tau$}
\State fullpop\_sim = cosine\_similarity($X$)
\State filtered\_sim = where(fullpop\_sim > $\tau$)
\State sampled\_edges = \textbf{random\_sample}(filtered\_sim) 
\State $S^{(0)}$ = graph(edges = edge\_idx, nodes = $\mathcal{G}$.nodes) 
\State \textbf{return} $S^{(0)}$
\EndProcedure
\end{algorithmic}
\end{algorithm}

\subsection{Statistics on Graph Datasets}\label{ap:distributions}
\begin{figure}[h!]
\vspace{-1em}
    \centering
    \includegraphics[width=0.85\linewidth]{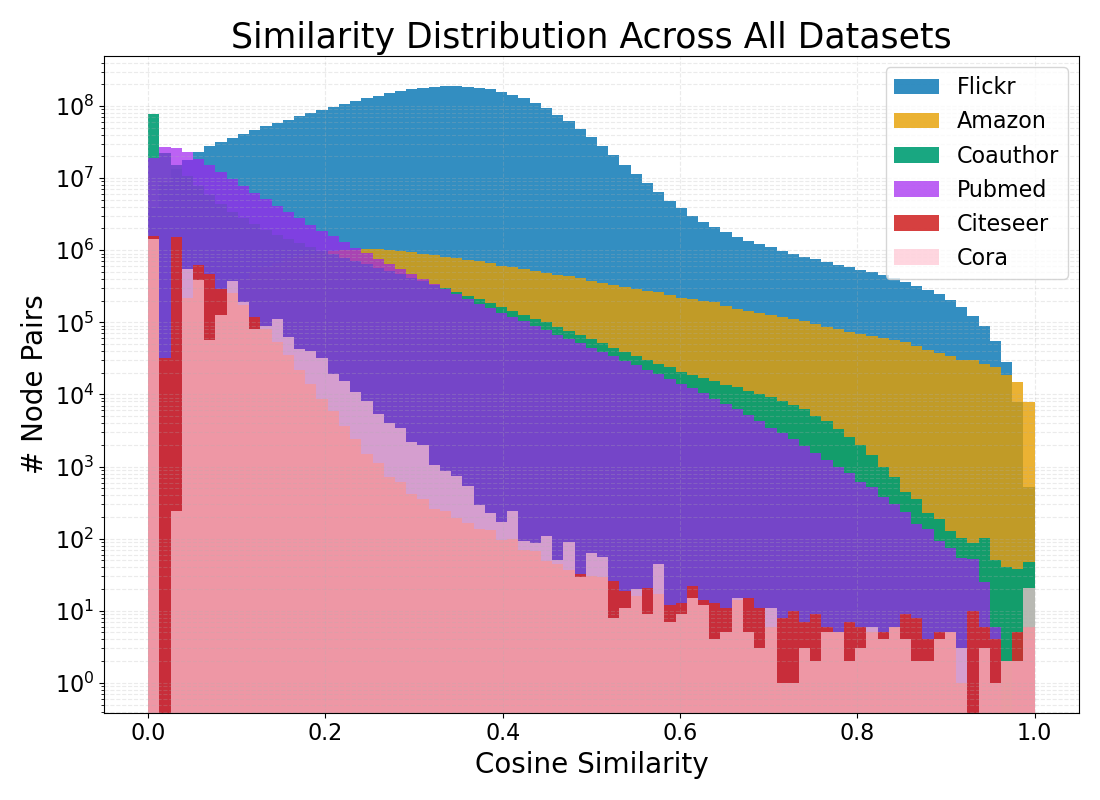}
     %\vspace{-1em}
    \caption{Distributions of pairwise cosine similarities $S_{ij}$.}% across the six graph datasets from our experiments.}
    \label{fig:distributions}
    \vspace{-2em}
\end{figure}

We present the key graph dataset statistics for contextualizing our results. In Table \ref{tab:dataset} we can see that our datasets range in size, density, class distribution, and feature dimension. In Figure \ref{fig:distributions} we highlight differences in distributions of pairwise similarities. 
\begin{table}[h!]
%\vspace{-1em}
     \centering
         \caption{Summary statistics of the graph datasets.}
         \vspace{-1em}
     \resizebox{\linewidth}{!}{
     \begin{tabular}{c|ccccc}
     \toprule
\textbf{Dataset}&\textbf{Type}&\textbf{Nodes}&\textbf{Edges}&\textbf{Classes}&\textbf{Features}   \\
         \midrule
          \textbf{Coauthor}&Citation Graph & 18,333& 163,788&15 & 6,805 \\ 
          \textbf{Amazon}& Purchase Interactions & 7,650 & 491,722  &10 &767\\ 
          \textbf{Flickr}& Social Network & 89,250 &899,756 &7 &500\\ 
          \textbf{Pubmed}& Citation Network & 18,717 &88,651 &3 &500\\ 
          \textbf{Cora}& Citation Network & 2,708 &10,556 &7 &1,433\\ 
          \textbf{Citeseer}& Citation Network & 3,327 &9,228 &6 &3,703\\ 
          \bottomrule        
     \end{tabular}}
     \label{tab:dataset}
     \vspace{-1em}
 \end{table}

\subsection{Hyperparameters}\label{ap:hp}
\subsubsection{Backbone Settings}
We train the backbone using the Adam optimizer~\cite{kingma2014adam}. The  architecture is a two-layer GCN, as presented in \citet{kipf_semi_2016}, with an output dimension equal to the number of node classes in each dataset (see Table~\ref{tab:dataset}). At each FairExpand iteration, we train the GCN until the F1 score stabilizes on the validation set. For subsequent iterations ($k = 2, \dots, 15$), we fine-tune the model using the extended matrix $S^{(k)}$. Dataset-specific learning rates and hidden dimensions are reported in Table~\ref{table:hp_settings}.
\subsubsection{FairExpand Settings}
For details of our implementation, please see: \url{https://anonymous.4open.science/r/FairExpand-WWW-47BC/README.md}. In Phase 1, we set the fairness–utility trade-off parameter in Equation~\eqref{eq:L_NC} to $\lambda = 0.5$ so that both terms contribute equally. In Phase 2, when training the PULL link predictor, we follow the original implementation~\cite{kim_accurate_2024} and add a ratio of $0.05$, meaning that at each iteration, 5\% of the highest-scoring edges are added to the expected adjacency graph. For the $\varepsilon$-greedy exploration policy used in PULL, we use $\varepsilon = 0.2$ across all datasets. We set the total number of FairExpand iterations to $K = 15$ for all methods. For baselines that do not support graph augmentation, we simply train them for 15 additional epochs to ensure comparable training budgets. Table~\ref{table:hp_settings} lists the values of $|S^{(0)}|$, $m_{\text{add}}$, and $\tau$ used to obtain the FairExpand results. The choice of each $\tau$ is motivated by Table~\ref{table:tau}, which shows that some datasets contain very few highly similar node pairs; in such cases, a lower threshold is required to construct a sufficiently informative partial similarity matrix $S^{(0)}$.

\begin{table*}[t!]

\centering
\caption{Evaluation under feature separation. Results  shown as F1/Bias without standard deviations for compactness. \textit{Full} uses the entire feature set for building $S^{(0)}$ and for classification. \textit{Half} uses half of the features for $S^{(0)}$ and half for classification.}
\label{table:feat_drop}
\resizebox{\linewidth}{!}{
\begin{tabular}{c||cc||cc||cc||cc||cc||cc}
\toprule

\textbf{Model} &
\multicolumn{2}{c||}{\textbf{Coauthor}} &
\multicolumn{2}{c||}{\textbf{Amazon}} &
\multicolumn{2}{c||}{\textbf{Flickr}} &
\multicolumn{2}{c||}{\textbf{Pubmed}} &
\multicolumn{2}{c||}{\textbf{Cora}} &
\multicolumn{2}{c}{\textbf{Citeseer}} \\
\cmidrule(lr){2-3}
\cmidrule(lr){4-5}
\cmidrule(lr){6-7}
\cmidrule(lr){8-9}
\cmidrule(lr){10-11}
\cmidrule(lr){12-13}

& \textbf{Full} & \textbf{Half}
& \textbf{Full} & \textbf{Half}
& \textbf{Full} & \textbf{Half}
& \textbf{Full} & \textbf{Half}
& \textbf{Full} & \textbf{Half}
& \textbf{Full} & \textbf{Half} \\

\midrule
\textbf{GCN (Reference)} &
0.92 / 32623.71 & 0.92 / 27114.47 &
0.93 / 62277.01 & 0.93 / 65066.92 &
0.51 / 10505.58 & 0.51 / 12519.87 &
0.87 / 2093.56  & 0.83 / 2214.47 &
0.86 / 359.65   & 0.85 / 210.19 &
0.76 / 111.46   & 0.75 / 100.97 \\

\midrule
\textbf{PFR} &
0.46 / 20.87  & 0.44 / 20.22 &
0.39 / 33.44  & 0.38 / 27.81 &
0.44 / 26.23  & 0.43 / 24.25 &
0.55 / 1.73   & 0.53 / 1.28 &
0.37 / 0.00   & 0.37 / 0.00 &
0.24 / 0.00   & 0.21 / 0.00 \\

\midrule
\textbf{InFoRM} &
0.88 / 8021.25 & 0.92 / 13400.14 &
0.85 / 8981.56 & 0.92 / 12627.11 &
0.38 / 905.64  & 0.51 / 11815.02 &
0.86 / 1526.11 & 0.83 / 2109.85 &
0.85 / 238.85  & 0.85 / 173.02 &
0.75 / 126.05  & 0.75 / 110.05 \\

\midrule
\textbf{REDRESS} &
0.32 / 23953.98 & 0.27 / 5482.98 &
0.24 / 22013.31 & 0.21 / 2637.91 &
0.43 / 1800746.47 & 0.33 / 199451.94 &
0.44 / 2.04    & 0.42 / 1.34 &
0.37 / 0.24    & 0.34 / 0.05 &
0.47 / 5.65    & 0.39 / 2.61 \\

\midrule
\textbf{GFairHint} &
0.91 / 657745.01 & 0.91 / 9.11e6 &
0.66 / 956512.86 & 0.89 / 271095.01 &
0.27 / 39525.90  & 0.44 / 17327.29 &
0.85 / 39525.90  & 0.83 / 16245.30 &
0.82 / 5412.02   & 0.82 / 11688.98 &
0.69 / 4219.85   & 0.66 / 6731.22 \\

\midrule
\textbf{FairExpand} &
0.78 / 2779.55 & 0.86 / 4022.75 &
0.81 / 5712.18 & 0.81 / 6994.70 &
0.42 / 40.66   & 0.42 / 931.55 &
0.86 / 1425.01 & 0.83 / 1105.64 &
0.85 / 261.01  & 0.84 / 136.02 &
0.76 / 76.92   & 0.73 / 92.20 \\

\bottomrule
\end{tabular}
}

\end{table*}

\begin{figure}[h!]
    \centering
    \includegraphics[width=.44\linewidth, trim={0cm 0cm 0cm 0},clip]{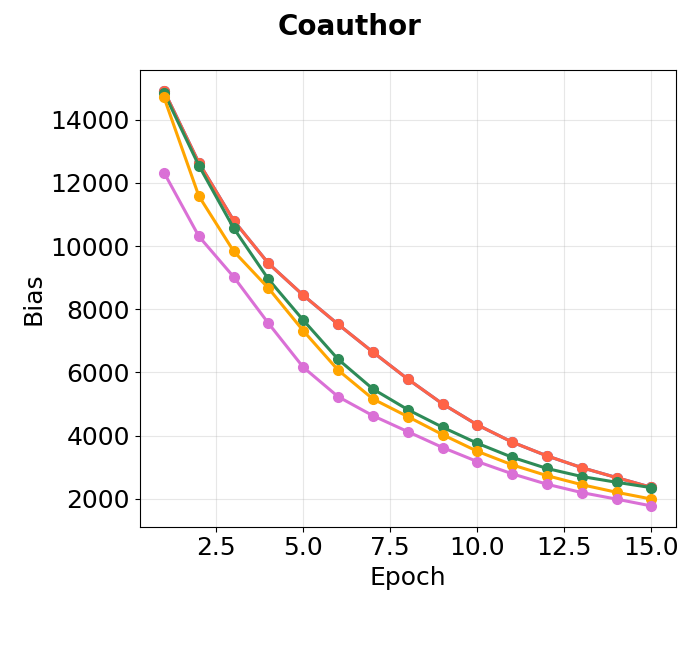}    
    \includegraphics[width=.44\linewidth, trim={0cm 0cm 0cm 0},clip]{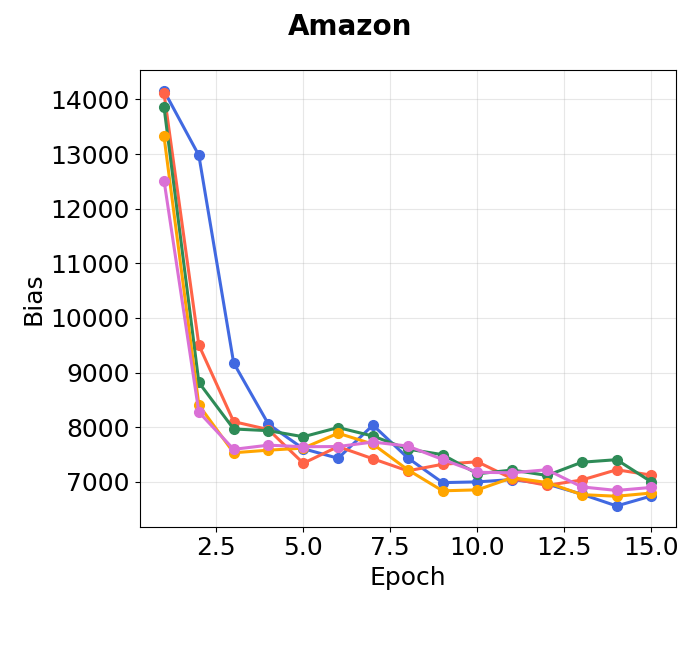}
    \includegraphics[width=.44\linewidth, trim={0cm 0cm 0cm 0},clip]{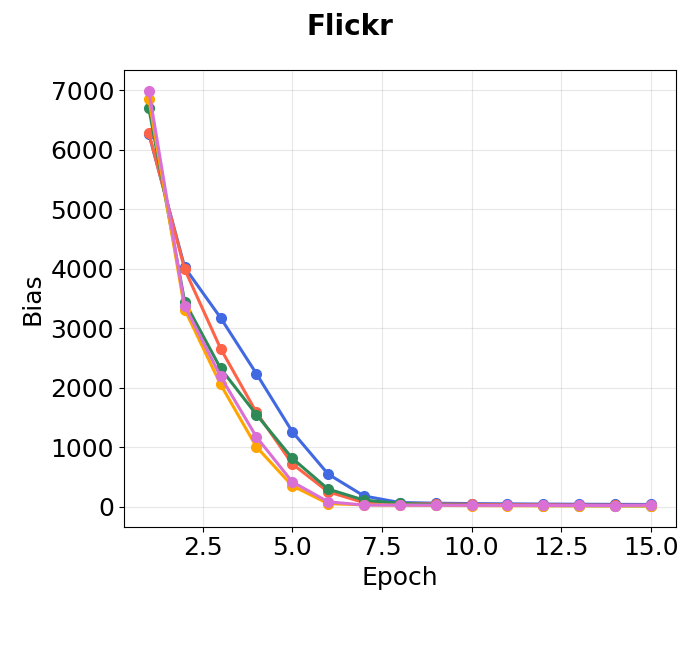}    
    \includegraphics[width=.44\linewidth, trim={0cm 0cm 0cm 0},clip]{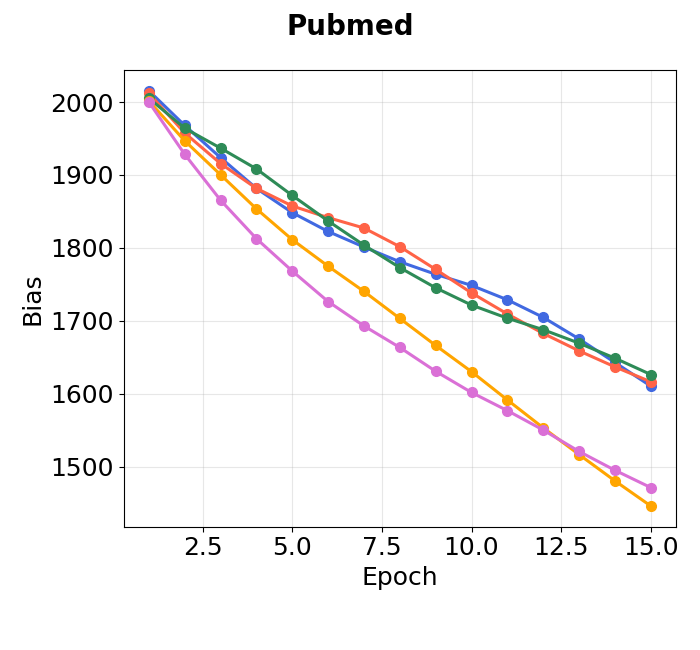}
    \includegraphics[width=.44\linewidth, trim={0cm 0cm 0cm 0},clip]{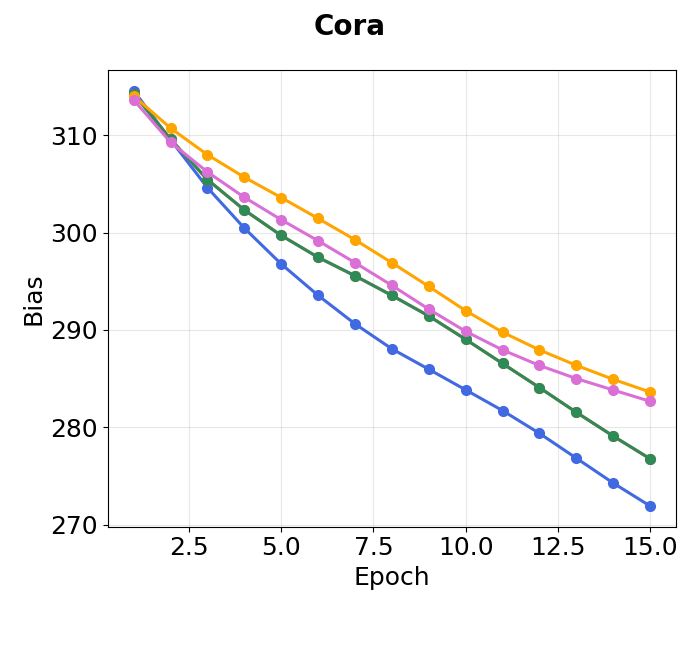}    
    \includegraphics[width=.44\linewidth, trim={0cm 0cm 0cm 0},clip]{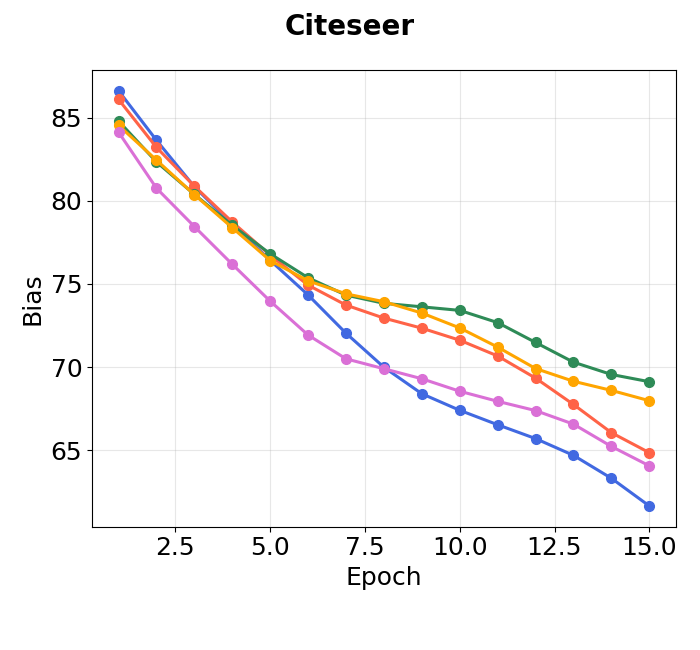}
    \includegraphics[width=\linewidth, trim={0cm 0cm 0cm 0},clip]{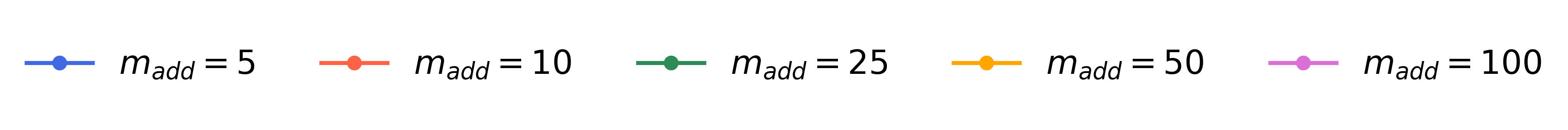}
    \caption{Bias trajectories over FairExpand iterations.} %With each step of the iterative expansion process, bias scores continue to decay asymptotically.}
    \label{fig:conv}
\end{figure}

\subsubsection{Baselines} 
All implementations can be found in the \texttt{baselines} folder in our repository. For all the methods listed below, we replace the dense similarity matrix $S$ with our sparse input $S^{(0)}$. For {PFR}\footnote{\url{https://github.com/plahoti-lgtm/PairwiseFairRepresentations}} \cite{lahoti2019operationalizing}, we use the original implementation and the authors' hyperparameters. For {InFoRM}\footnote{\url{https://github.com/jiank2/InFoRM}} \cite{kang_inform_2020} and {REDRESS}\footnote{\url{https://github.com/yushundong/REDRESS}} \cite{dong_individual_2021}, we integrate their fairness frameworks into our training pipeline. Since they share the same architecture as Phase~1 of FairExpand's training paradigm, we apply the same tuned hyperparameters for these methods. For {GFairHint}\footnote{\url{https://github.com/paihengxu/GFairHint}} \cite{xu_gfairhint_2023}, we use the original implementation, modifying only the evaluation metrics. We use the GCN+InFoRM version of their framework to ensure the most accurate comparison.

\begin{table}[t]
\centering
\begin{minipage}{\linewidth}
\centering
\caption{Hyperparameter settings of FairExpand in Table \ref{table:overall_performance}.}
\vspace{-1em}
\label{table:hp_settings}
\resizebox{0.9\linewidth}{!}{
\begin{tabular}{c|ccc|cc}
\toprule
\textbf{Dataset} 
& $|S^{(0)}|$ & $m_{\text{add}}$ & $\tau$ 
& \textbf{Learning Rate} & \textbf{Hidden Dim} \\
\midrule
Coauthor & 50 & 10 & 0.7 & 0.01  & 32  \\
Amazon   & 50 & 10 & 0.9 & 0.005 & 64  \\
Flickr   & 20 & 10 & 0.7 & 0.01  & 128 \\
Pubmed   & 20 & 10 & 0.7 & 0.01  & 16  \\
Cora     & 20 & 10 & 0.4 & 0.005 & 64  \\
Citeseer & 20 & 10 & 0.4 & 0.01  & 64  \\
\bottomrule
\end{tabular}
}
\end{minipage}
\vspace{0.25cm}
%\hfill

\begin{minipage}{\linewidth}
\centering
\caption{Number of similar training pairs across $\tau$ thresholds.}\label{table:tau}
\vspace{-1em}
\resizebox{\linewidth}{!}{
\begin{tabular}{c|cccccc}
\toprule
\textbf{Dataset} & $\tau{=}0.4$ & $\tau{=}0.5$ & $\tau{=}0.6$ & $\tau{=}0.7$ & $\tau{=}0.8$ & $\tau{=}0.9$ \\
\midrule
Coauthor & 195{,}870 & 68{,}628 & 24{,}721 & 7{,}811 & 1{,}216 & 93 \\
Amazon   & 1{,}394{,}373 & 800{,}304 & 427{,}810 & 206{,}029 & 85{,}540 & 23{,}923 \\
Flickr   & 140{,}513{,}325 & 22{,}542{,}843 & 4{,}269{,}595 & 1{,}646{,}091 & 612{,}865 & 107{,}521 \\
Pubmed   & 146{,}257 & 45{,}262 & 12{,}408 & 2{,}587 & 310 & 31 \\
Cora     & 177 & 60 & 26 & 17 & 14 & 7 \\
Citeseer & 137 & 63 & 43 & 23 & 11 & 5 \\
\bottomrule
\end{tabular}
}
\end{minipage}
%\vspace{-1em}
\end{table}

\section{Additional Experimental Results}

\subsection{Bias Trajectories of FairExpand}\label{sec:conv}
Figure \ref{fig:conv} showcases the individual bias trajectories in FairExpand for iterations $k=1, \dots, K$. The results show that each FairExpand iteration consistently reduces bias across all datasets, thereby improving individual fairness. We also illustrate the differences obtained by changing the number of edges $m_{\text{add}}$ added at each iteration.

\subsection{Evaluation Under Feature Separation}\label{sec:extra}
Table~\ref{table:overall_performance} follows the evaluation protocol of prior work~\cite{kang_inform_2020, dong_individual_2021, xu_gfairhint_2023}. Here, we conduct another controlled analysis where half of the feature set is used to define similarities and the other half is used to train models. 
As shown in Table~\ref{table:feat_drop}, even under this more challenging setting designed to prevent any leakage from features, FairExpand continues to outperform baselines in bias mitigation. This confirms that the strong performance observed in Table~\ref{table:overall_performance} is intrinsic to our approach rather than the result of any potential feature leakage.

\end{document}